\documentclass[sigconf]{aamas} 

\usepackage{graphicx}
\usepackage{subcaption}
\usepackage{xspace}

\newcommand{\ourname}{Heuristic Transformer\xspace}
\newcommand{\acronym}{HT\xspace}

\usepackage{balance}        

\usepackage{algorithm}
\usepackage{algorithmic}

\title{Heuristic Transformer: Belief Augmented In-Context Reinforcement Learning}

\author{Oliver Dippel}
\affiliation{
  \institution{University of Liverpool}
  \city{Liverpool}
  \country{United Kingdom}}
\email{oliver.dippel@liverpool.ac.uk}

\author{Alexei Lisitsa}
\affiliation{
  \institution{University of Liverpool}
  \city{Liverpool}
  \country{United Kingdom}}
\email{A.Lisitsa@liverpool.ac.uk}

\author{Bei Peng}
\affiliation{
  \institution{University of Sheffield}
  \city{Sheffield}
  \country{United Kingdom}}
\email{bei.peng@sheffield.ac.uk}

\begin{abstract}
Transformers have demonstrated exceptional in-context learning (ICL) capabilities, enabling applications across natural language processing, computer vision, and sequential decision-making. In reinforcement learning, ICL reframes learning as a supervised problem, facilitating task adaptation without parameter updates. Building on prior work leveraging transformers for sequential decision-making, we propose Heuristic Transformer (HT), an in-context reinforcement learning (ICRL) approach that augments the in-context dataset with a belief distribution over rewards to achieve better decision-making. Using a variational auto-encoder (VAE), a low-dimensional stochastic variable is learned to represent the posterior distribution over rewards, which is incorporated alongside an in-context dataset and query states as prompt to the transformer policy. We assess the performance of HT across the Darkroom, Miniworld, and MuJoCo environments, showing that it consistently surpasses comparable baselines in terms of both effectiveness and generalization. Our method presents a promising direction to bridge the gap between belief-based augmentations and transformer-based decision-making.
\end{abstract}

\keywords{Transformer Model, In-Context Reinforcement Learning, Bayes-adaptive Markov Decision Processes}

\newcommand{\BibTeX}{\rm B\kern-.05em{\sc i\kern-.025em b}\kern-.08em\TeX}

\renewcommand\footnotetextcopyrightpermission[1]{}  
\makeatletter
\def\@copyrightspace{}  
\makeatother

\begin{document}

\pagestyle{fancy}
\fancyhead{}

\maketitle 



\section{Introduction}

The transformer model, introduced by \citet{Attention}, has become a cornerstone in various domains of machine learning, such as natural language processing \citep{radford2018improving, devlin2018bert, dai2019transformer}, computer vision \citep{dosovitskiy2020image}, and sequential decision-making \citep{chen2021decision, janner2021offline}. A key capability of transformers is in-context learning (ICL) \citep{brown2020language, liu2023pre}. In reinforcement learning (RL), this ability reduces learning to a supervised problem and enables tackling new tasks in context without parameter updates following a pre-training phase.

\begin{figure*}[t]
  \centering
  \includegraphics[width=0.9\textwidth]{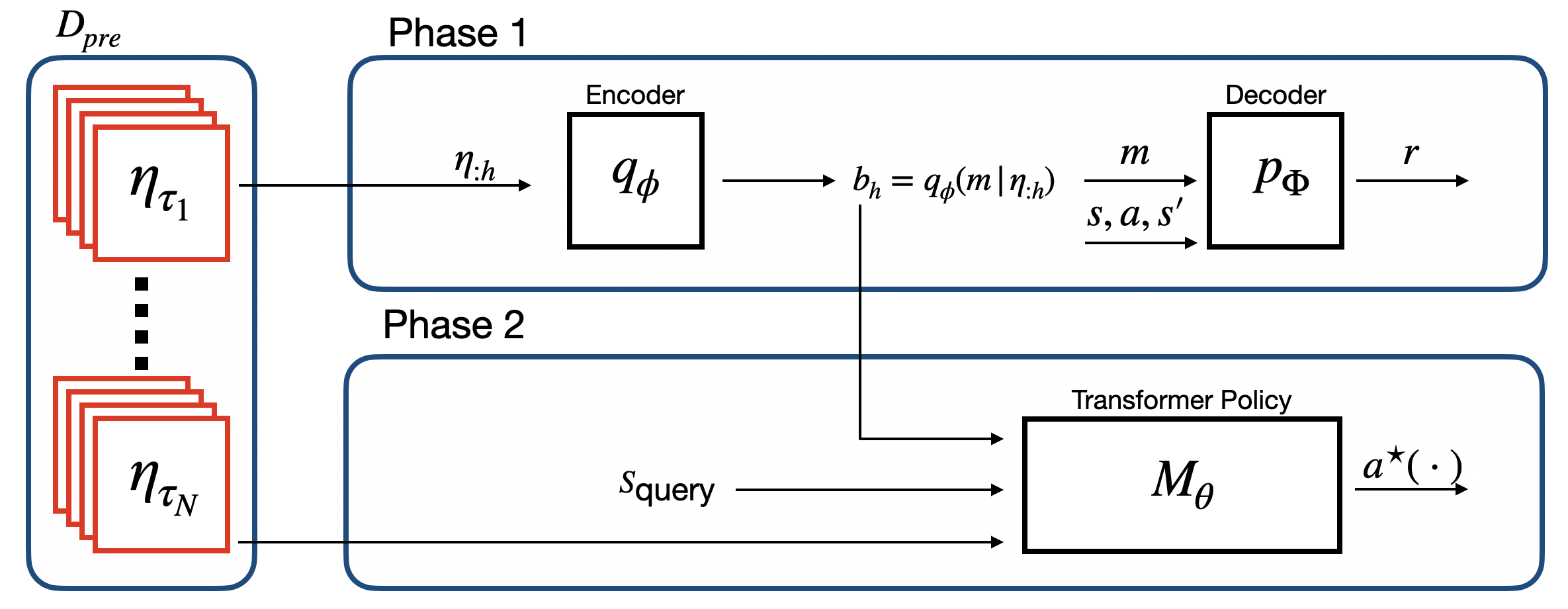}
  \caption{Illustration of the two-phase framework in \acronym. In Phase 1, a variational autoencoder (VAE) infers the belief distribution over reward functions $b_{h}=q_{\phi}(m\mid\eta_{:h})$ from offline transitions $\eta$. In Phase 2, the learned belief $b_h$, combined with a query state $s_{\text{query}}$ and the in-context dataset $D_{\text{pre}}$ form the input to the transformer policy model $M_{\theta}$ to learn the optimal action distribution $a^{\star}(\cdot)$.}
  \label{fig:model}
\end{figure*}

Recent research has successfully employed the ICL abilities of transformers in sequential decision-making problems \citep{laskin2022context, sinii2023context, zisman2023emergence, lee2024supervised, dippel2024contextual, huang2024context}. Although, mostly infeasible for online learning \citep{online_DT}, these models are capable of predicting the next action based on a query state and a history of environment interactions, which inform the learner about task objectives and environment dynamics. While methods like Algorithm Distillation (AD) \citep{laskin2022context} focus on learning policy improvements of source algorithms, \citet{lee2024supervised} propose Decision-Pretrained Transformer (DPT), which learns the rewards and environment dynamics of unsorted source trajectories and is able to perform posterior sampling under certain conditions. Specifically, an agent samples step-wise from its posterior and acts optimally according to the sampled Markov decision process (MDP) until the next step is done and the posterior is updated. Posterior sampling is often performed as a common approximation to a Bayes-optimal policy \citep{thompson1933likelihood, strens2000bayesian, osband2013more}, for environments where such policies are intractable. However, sampling from a posterior can be highly inefficient and diverge from Bayes-optimal. 

A Bayes-optimal policy can theoretically be computed using a framework of Bayes-adaptive Markov decision processes (BAMDPs) \citep{martin1967bayesian, duff2002optimal}, where the agent maintains a belief distribution over possible MDPs within the environment. 
However, learning a Bayes-optimal policy is unfeasible in most environments due to computational intractability.
Prior work in meta-RL, such as VariBAD \citep{varibad}, operating in the framework of BAMDPs, introduces the concept of augmenting the agent's state with a belief distribution over possible MDPs. By doing so, VariBAD is able to approximate Bayes-optimal behaviour. In VariBAD, a variational auto-encoder (VAE) is used to infer the belief based on observed transitions, allowing the agent to adapt its policy dynamically. This framework has shown success in improving task generalization through belief augmentation. 

DPT and VariBAD represent two distinct yet complementary paradigms in decision-making. DPT leverages the ICL capabilities of transformers to predict actions directly from historical interactions, whereas VariBAD explicitly models task uncertainty through belief distributions, augmenting the agent’s state to approximate Bayes-optimal behavior. In our work, we aim to develop a hybrid approach that combines the generalization benefits of belief augmentation with the efficiency and scalability of ICL-enabled transformer architectures in RL.

Recent work by \citet{dippel2024contextual} proposes Goal-Focused Transformer (GFT), which extends DPT by implicitly incorporating task-specific information into the in-context dataset after observing the goal location post hoc, thereby providing useful information about the unknown reward function. While GFT improves upon DPT’s performance for certain environments, it relies heavily on extensive meta-knowledge of the environment in order to construct the context prompt as an indirect representation of the reward function. This dependency limits GFT’s applicability across diverse environments but underscores the potential value of incorporating additional information, such as a belief distribution, to improve in-context reinforcement learning (ICRL) methods.

Building upon these insights, we propose a novel ICRL method called \ourname (\acronym), which combines the fundamental idea of BAMDPs by modelling a belief distribution over rewards with the in-context abilities of transformer models.
Inspired by VariBAD, we use a VAE to learn a low-dimensional stochastic variable, $m$, which represents the posterior belief of the reward function. Unlike VariBAD, where $m$ encodes the posterior belief of the underlying MDP, our method operates in an offline ICRL setting, utilizing pre-training without requiring online interaction during training. Compared to GFT, our approach generalizes more effectively across tasks by incorporating a learned belief distribution that does not rely on external meta-knowledge. By passing this belief as an additional prompt to the transformer policy, we enable the agent to better handle task uncertainty and achieve superior performance and higher expected returns during online evaluation, surpassing both DPT and GFT. 

To train \acronym, we split the training process into two sequential phases. In the first phase, a VAE infers the posterior distribution of rewards, referred to as the belief $b$, for a new task, given past $<$state, action, reward, next state$>$ transitions. In the second phase, the belief $b$, together with an in-context dataset $D_{\text{pre}}$ consisting of the same past transitions and a query state $s_{\text{query}}$, forms the prompt to the transformer model to predict the optimal action. The network architecture is summarized in Figure \ref{fig:model}. During online evaluation, the agent's experience from the last rollouts along with the belief $b$ updates the agent's behaviour completely in context, without any parameter updates. 

This paper proposes \ourname (\acronym), a new ICRL method that combines the fundamental idea of BAMDPs by modelling a belief distribution over rewards with the in-context abilities of transformer models. 
Our experimental results demonstrate that HT outperforms baseline ICRL methods in various domains featuring both discrete and continuous action spaces, in terms of online adaptation and generalization.

\section{Background}

The foundational decision-making framework for this study is the finite-horizon MDP. An MDP is defined by the tuple $\tau = (S, A, T, R, H, \allowbreak \omega, \gamma)$. We define a task $\tau$  as a specification of an MDP, where $S$ represents the state space, $A$ denotes the action space, $T:S \times A \rightarrow S$ is the transition function, $R:S \times A \rightarrow R$ specifies the reward function, $H\in \mathbb{N}$ indicates the time horizon, $\omega \in (S)$ is the initial state distribution and $\gamma$ being a discount factor. The interaction between the learner and the environment follows a defined protocol. First, an initial state $s_1$ is sampled from $\tau$. Sampling an MDP from $\tau_{i} \sim p(\tau)$ is typically done by sampling a reward and transition function from a distribution $\rho(R, T)$. At each subsequent time step $h$, the learner selects an action $a_h$, causing a transition to a new state $s_{h+1} \sim T(\cdot|s_h, a_h)$ and receives a reward of $r_{h+1} \sim R(\cdot |s_h, a_h)$. This process continues until the episode concludes after $H$ steps. A policy $\pi$ maps states to distributions over actions and determines the learner’s interactions with the MDP. The optimal policy, denoted as $\pi^{*}$, maximizes the value function $V(\pi^{*}) := max_{\pi}\mathbb{E}_{\pi}\sum_{h=1}^H[ \gamma^{h}r_{h}]$, with $\gamma$ being a discount factor.

To simplify the framework, we assume the state space is partitioned by $h \in [H]$, allowing the optimal policy $\pi^{*}$ to remain notationally independent of $h$. With that in mind, $h$ differs from the common notation $t$, indicating time independence during training, i.e., the training dataset is a collection of transition tuples generated within $\tau$ in an unsorted order.

\subsubsection{Heuristic Bayesian Policy}

When the underlying MDP is unkown, optimal decision-making necessitates a balance between exploration and exploitation during action selection. A principled way to address this challenge is through a Bayesian approach to reinforcement learning, which can be formalized as a BAMDP. The solution to this framework is a Bayes-optimal policy, as outlined in foundational works \citep{bellman1956problem, duff2002optimal, ghavamzadeh2015bayesian}. 

In Bayesian reinforcement learning, the agent operates in an environment where the reward and transition dynamics are initially unknown and modeled as random variables. These variables are characterized by a prior distribution, denoted as $b_{0} = \rho(R)$. This formalism enables the agent to leverage a probabilistic framework, continuously updating its understanding of the environment as it gathers experience. At each step $h$, the agent maintains a belief distribution $b_{h} = \rho(R) = \rho(R|\eta_{:h})$, representing the posterior over the rewards given its accumulated trajectory $\eta_{:h} = \{(s, a, r, s')_{1}, \dots, (s, a, r, s')_{h}\}$. This Bayesian posterior reflects the integration of prior knowledge with observed evidence. Note that in this work, we make a deliberate simplification: rather than maintaining a full joint belief over both the reward function and the transition dynamics, we restrict our focus to modeling a posterior belief over the rewards alone. This choice significantly specializes on the reward function, but at the cost of no longer achieving full Bayes-optimality. Without a complete posterior that incorporates the transition dynamics, the policy derived from this belief cannot fully adhere to the principles of Bayesian decision theory. Nevertheless, the resulting policy leverages the Bayesian framework heuristically to guide decision-making under uncertainty. By focusing on the reward dynamics, we designate the resulting strategy as a Heuristic Bayesian Policy. This name reflects its roots in Bayesian reasoning while acknowledging the specialization of this design.

\section{Heuristic Transformer}\label{sec:mod}

In-context reinforcement learning methods usually go through a pre-training phase in which the transformer policy learns to predict actions in context. Our method focuses on learning two key components: (1) a compact representation of the reward function dynamics and (2) a policy model capable of predicting optimal actions given a belief over the reward function, an in-context dataset, and a query state. In the first phase, the objective is to infer a posterior belief distribution over the reward functions of tasks, encapsulated in a latent embedding $m$. This embedding compresses the reward dynamics into a low-dimensional representation that can be efficiently used by the policy model. The training optimizes a variational Evidence Lower Bound (ELBO), which enables tractable posterior inference through a learned variational distribution. In the second phase, the policy model is trained using supervised learning to map in-context datasets, the belief over the reward function, and query states to optimal actions. This process conditions the model to generalize across tasks sampled from a pre-training task distribution. The resulting model, referred to as \ourname, can predict optimal actions in a variety of contexts by leveraging its learned representations of task-specific reward dynamics and policies.

\subsubsection*{Phase 1}
Let $T_\text{{pre}}$ be a distribution over tasks $\tau$ during pre-training. The distribution $T_\text{{pre}}$ can span different reward and transition functions, as well as different state and action spaces. To begin with, we sample a prompt consisting of a dataset $D \sim D_\text{{pre}}(\cdot;\tau)$ of interactions between the learner and the MDP specified by $\tau$. The dataset $D = \{(s_{j}, a_{j}, s'_{j}, r_{j} )\}_{j \in [n]}$ is a collection of transition tuples generated within $\tau$. This dataset $D$, referred to as the in-context dataset, provides contextual information about $\tau$. The dataset $D$ can be generated through various methods, including random interactions within $\tau$, demonstrations from an expert and rollouts of a source algorithm. In this work, we consider the actions to come from a random policy, $a \sim \pi_{\text{ran}}$.

In the first pre-training phase, we aim to learn the posterior belief of the reward distribution. Accurately computing the posterior is generally infeasible due to the lack of access to the MDP, including the transition and reward functions, as well as the computational challenges. To address this, we develop a parameterized model of the reward, denoted as $p_{\Phi}(r_{:H}|m, \eta_{:H-1})$, and an inference network,  $q_{\phi}(m|\eta_{:H})$. The latter enables efficient inference at each step $h$. Here, $\Phi$ and $\phi$ represent the parameters of the reward model and inference network, respectively, and $m$ represents the embedding. Our goal is to learn a belief distribution over the reward functions in $\tau$, using posterior knowledge of the environment's internal reward function, to compute the optimal action. With the above formulation, it becomes sufficient to reason about the embedding $m$ rather than the full reward dynamics. This approach is particularly advantageous for compressing the reward function dynamics into a compact vector $m$, which provides dense information to our policy model $M_{\theta}$.

The action selection does not form part of the MDP, so an environment model can only generate trajectory distributions when conditioned on actions, which are typically sampled from a policy. In our case, the action selection comes from a random policy $a \sim \pi_{\text{ran}}$, which is included in $\eta$. At any step h, the objective for learning the model is to maximize:

\begin{eqnarray}\label{eq:1}
    \mathbb{E}_{\rho_{\tau}(D)} \left[ \log p_{\Phi}(r_{:h}|\eta_{h-1}) \right],
\end{eqnarray} 

with $\rho_{\tau}(D)$ denoting the dataset distribution depending on $\tau$. Instead of directly optimizing the intractable objective of (\ref{eq:1}), we can target a tractable lower bound using a learned approximate posterior $q_{\phi}(m|\eta_{:h})$, which can be estimated via Monte Carlo sampling. This lower bound, derived in detail in Appendix A, is expressed as:

\begin{eqnarray}\label{eq:2}
\begin{split}
        \mathbb{E}_{\rho_{\tau}(D)} [\log p_\Phi (r_{:h})] \geq \\ 
        \mathbb{E}_{\rho_{\tau}} [ \mathbb{E}_{q_\phi (m \mid \eta_{:h})} [\log p_\Phi (r_{:h}|m)] 
- \text{KL}(q_\phi (m|\eta_{:h}) \| p_\Phi (m)) \\
=\mathrm{ELBO}_h.
\end{split}
\end{eqnarray}

This is known as the Evidence Lower Bound (ELBO), denoted as $\mathrm{ELBO}_h$. The term $\mathbb{E}_{q} [\log p(r_{:h}|m)]$, often called the reconstruction loss, is associated with the decoder $p(r_{:h}|m)$. The KL divergence $\text{KL}(q_\phi (m|\eta_{:h}) \| p_\Phi (m))$ quantifies the difference between the variational posterior $q_{\phi}$ and the prior over embeddings $p_{\Phi}(m)$. In our setup, the prior is defined recursively as the previous posterior, i.e., $p_\Phi(m)=q_\phi(m \mid \eta_{:h-1})$, with the initial prior set to $p_\Phi(m) = \mathcal{N}(0, I)$.

As depicted in Equation \ref{eq:2}, at timestep $h$, the past trajectory $\eta_{:h}$ is encoded to derive the current posterior $q(m|\eta_{:h})$, leveraging all available information to infer the reward. The model then decodes the full reward trajectory $r_{:h}$ to compute $\mathbb{E}_{q} [\log p(r_{:h}| m)]$. The reconstruction term $\log p(r_{:h}|m)$ can be factorized as:

\begin{eqnarray}
\begin{split}
    \log p(r_{:h}|m) = \\ \sum_{i=0}^{h-1} \left[\log p(r_{i+1} \mid s_i, a_i, s'_{i}; m)\right],
\end{split}
\end{eqnarray}

with $p(r_{i+1}|s_i , a_i, s'_{i}, m)$ representing the reward function. The so learned posterior distribution $q(m|\eta_{:h})$, which models the reward dynamics in the latent space $m$, encapsulates the belief $b_{h} = q(m|\eta_{:h})$. The belief $b$ is subsequently used by the policy model $M_{\theta}$.  

\subsubsection*{Phase 2}
In the second pre-training phase, we aim to learn an optimal policy model conditioned on the in-context dataset $D$, the belief $b$, and an additional query state $s_{\text{query}}$, which is independently sampled from the state space $S$. A corresponding action label $a^{*}$ is sampled from the optimal policy $\pi^{*}_{\tau} (\cdot | s_{\text{query}})$ for task $\tau$. The joint pre-training distribution over tasks, in-context datasets, beliefs, query states, and action labels is denoted as:

\begin{eqnarray}
\begin{split}
          \rho_{\text{pre}}(\tau, D, b, s_{\text{query}}, a^{*}) = \\ T_{\text{pre}}(\tau) D_{\text{pre}}(D; \tau) q(m|D, \tau)
    D_{\text{query}}(s_{\text{query}}) \pi^{*}_{\tau}(a^{*}| s_{\text{query}}).
\end{split}
\end{eqnarray}
   
Given the in-context dataset $D$, the belief $b$, and a query state $s_{\text{query}}$, a model can be trained to predict the optimal action $a^{*}$ through supervised learning. Let 
\begin{eqnarray*}
    D_j = \{(s_{1}, a_{1} , s'_{1}, r_{1}), \dots, (s_{j}, a_{j} , s'_{j}, r_{j}) \} 
\end{eqnarray*}

represent the partial dataset up to $j$ samples. We train a model $M$, parameterized by $\theta$, which outputs a distribution over actions $A$, by minimizing the expected loss over samples from the pre-training distribution:

\begin{eqnarray}\label{eq:3}
    \min_\theta \mathbb{E}_{\rho_{\text{pre}}} \mathbb{E}_{j \in [n]} \, \ell(M_\theta(\cdot| D_j, b_{j}, s_{\text{query}}), a^\ast).
\end{eqnarray}

This framework accommodates both discrete and continuous action spaces $A$. For experiments involving discrete $A$, we use a softmax parameterization for the output distribution of $M_{\theta}$, treating it as a classification problem. In the continuous case, the output of the transformer model is interpreted directly as the continuous action vector. The resulting trained model $M_{\theta}$ can be viewed as an algorithm that takes an in-context dataset of interactions $D$ and produces predictions of the optimal action for a belief $b$ and given query state $s_{\text{query}}$ via a forward pass. We refer to this trained model $M_{\theta}$ as \ourname.

\subsubsection{Training Objective}

Based on the previous formulations, we now summarize the training objectives for learning the approximate posterior distribution over reward embeddings, the generalized reward function, and the policy model. The objectives are trained sequentially starting with $\mathrm{ELBO}$:

\begin{eqnarray}\label{eq:4}
    \mathcal{L}_{1}(\Phi, \phi) = \max_{\Phi, \phi} \mathbb{E}_{D_{\text{pre}}}\sum_{j \in [n]}\mathrm{ELBO}_{j}(\Phi, \phi),
\end{eqnarray}

which can be optimized using the reparameterization trick \citep{kingma2013auto}, with the prior being Gaussian $q_{\phi}=\mathcal{N}(0, I)$. We encode trajectories using a Variational Autoencoder, similar to the architectures used in \citep{higgins2017darla, gregor2018temporal, hausknecht2015deep, hafner2019learning}. We now can incorporate the approximate posterior distribution over reward embeddings, i.e., the belief $b$, to train our policy model:

\begin{eqnarray}\label{eq:5}
    \mathcal{L}_{2}(\theta) = \min_{\theta} \mathbb{E}_{\rho_{\text{pre}}}\sum_{j \in [n]} -\log M_{\theta}(a^{*}|D_{j}, b_{j}, s_{\text{query}}),
\end{eqnarray}

this is equivalent to setting the loss function in (\ref{eq:3}) as the negative log-likelihood. We represent our policy as a causal Generative Pre-training Transformer (GPT) as in \citep{lee2024supervised, dippel2024contextual}.

\section{Related Work}


\begin{figure*}[t] 
  \centering

  \begin{subfigure}[t]{0.24\textwidth}
    \centering
    \includegraphics[width=\linewidth]{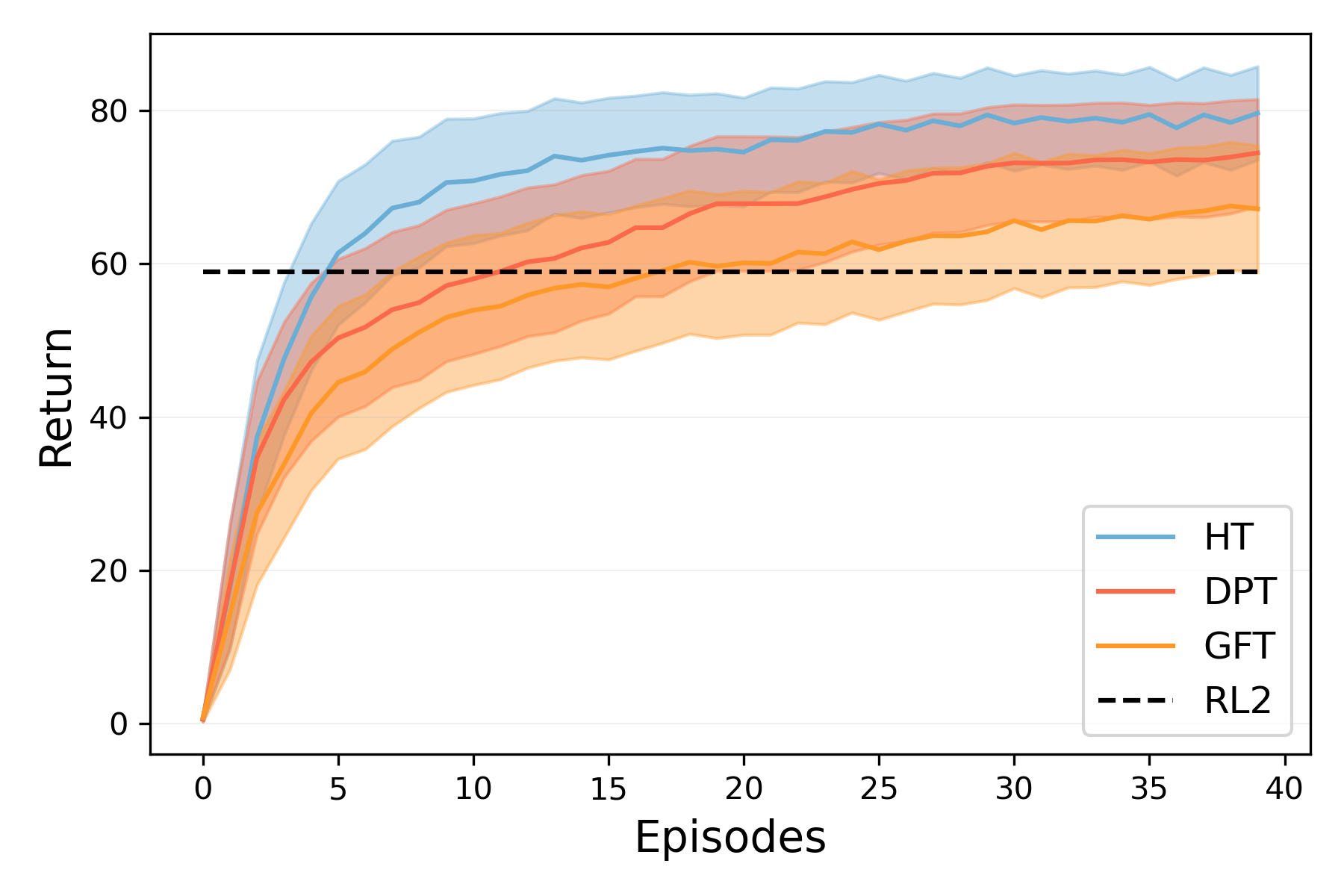}
    \caption{Darkroom}
    \label{fig:darkroom:a}
  \end{subfigure}\hfill
  \begin{subfigure}[t]{0.24\textwidth}
    \centering
    \includegraphics[width=\linewidth]{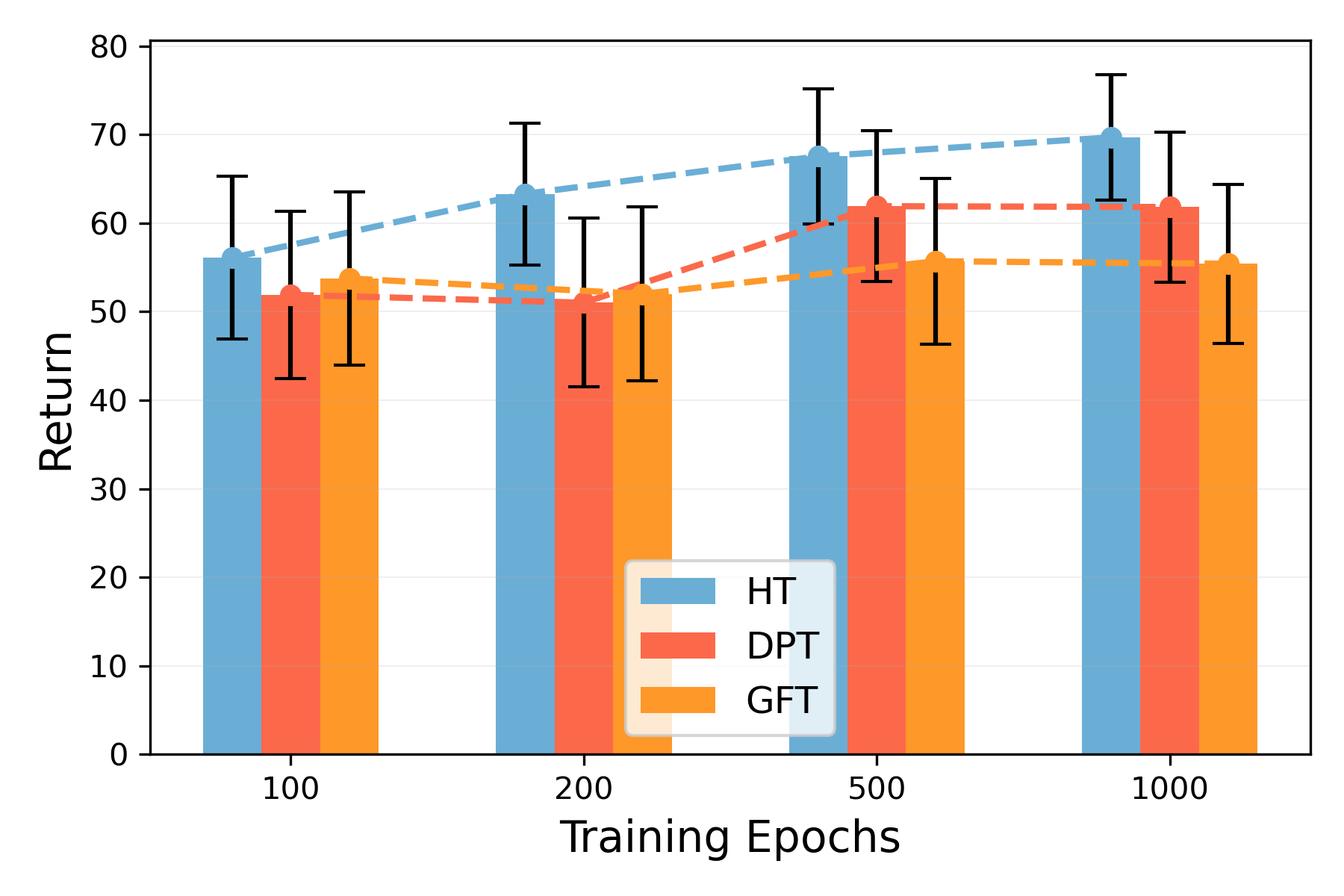}
    \caption{Darkroom (epochs)}
    \label{fig:darkroom:b}
  \end{subfigure}\hfill
  \begin{subfigure}[t]{0.24\textwidth}
    \centering
    \includegraphics[width=\linewidth]{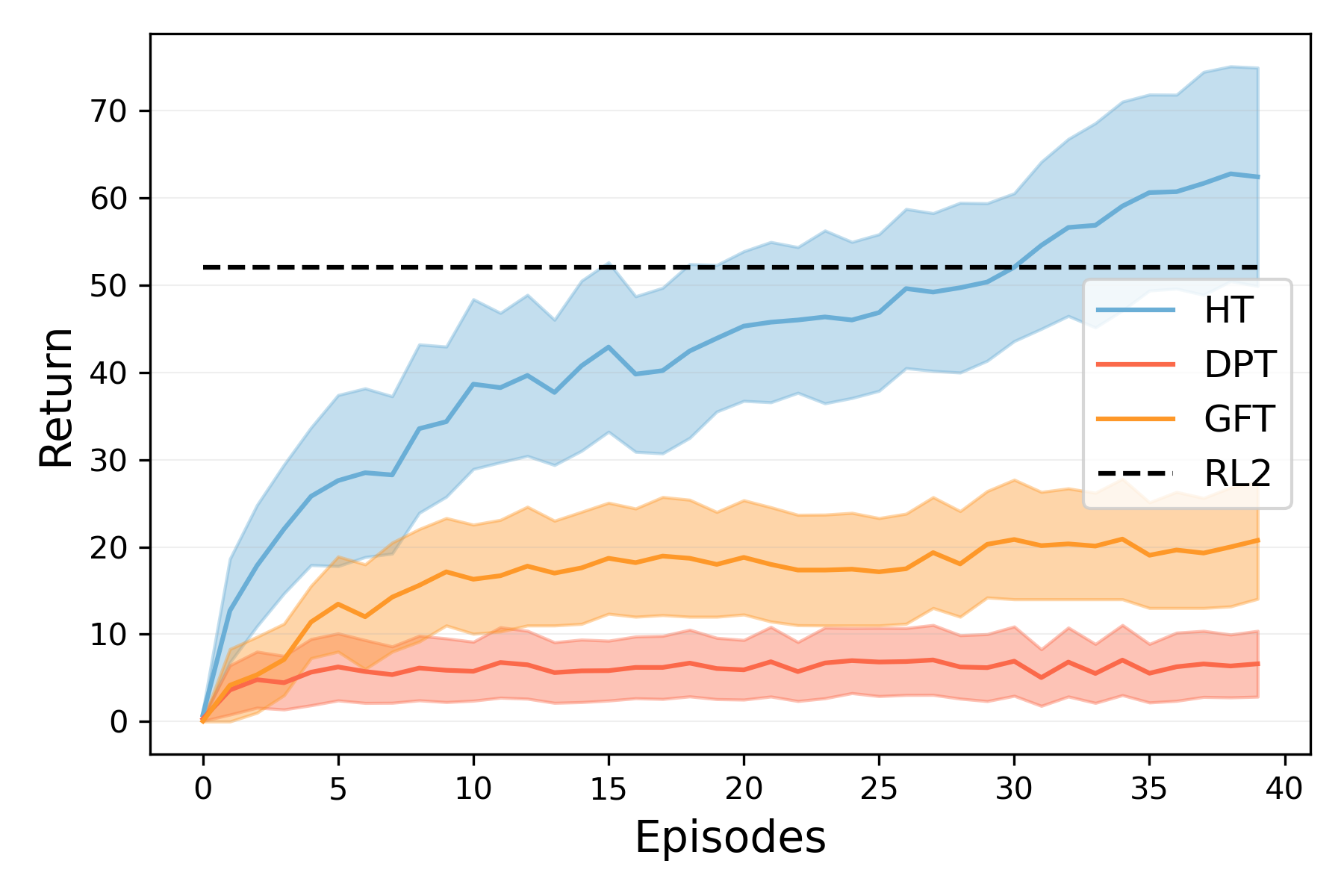}
    \caption{Darkroom Hard}
    \label{fig:darkroom:c}
  \end{subfigure}\hfill
  \begin{subfigure}[t]{0.24\textwidth}
    \centering
    \includegraphics[width=\linewidth]{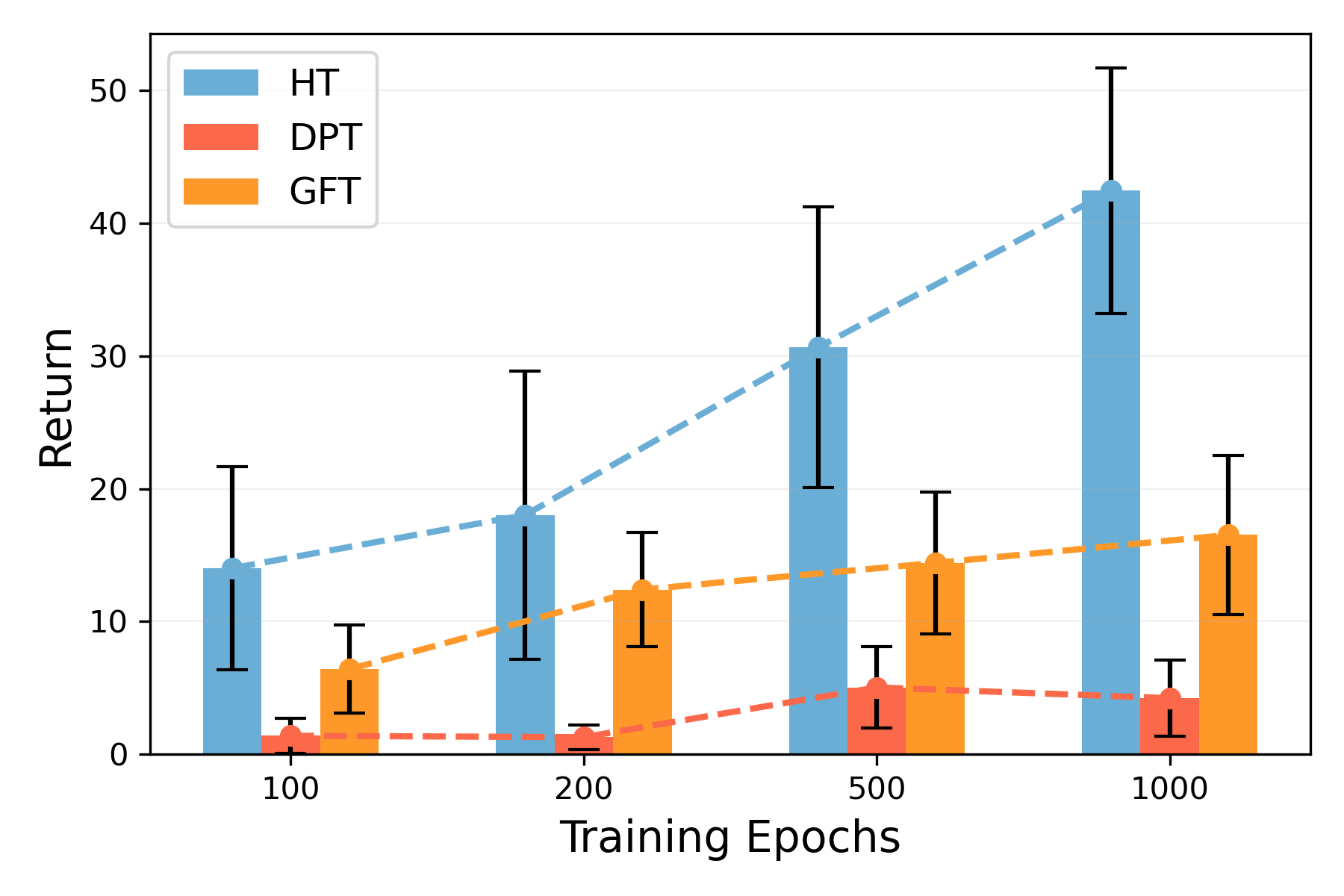}
    \caption{Darkroom Hard (epochs)}
    \label{fig:darkroom:d}
  \end{subfigure}

  \caption{(a) Online performance on test goals in Darkroom. (b) Darkroom after certain pre-training epochs. 
  (c) Darkroom Hard. (d) Darkroom Hard after certain pre-training epochs.
  Results are mean return $\pm$ std over 20 trials across 10 seeds.}
  \label{fig:darkroom}
\end{figure*}

Reinforcement learning (RL) traditionally operates as an online methodology, where agents learn by interacting with the environment in real time \citep{sutton2018reinforcement}. However, online RL poses challenges in scenarios where real-time interaction is impractical, costly, or unsafe. To address these limitations, offline RL allows agents to learn from a fixed dataset of pre-collected transitions, removing the need for ongoing interaction during training \citep{fujimoto2019off, kumar2020conservative, yu2021combo, kumar2019stabilizing}. Notable advancements in offline RL have explored transformer-based models for sequential decision-making by framing RL tasks as sequence modeling problems. For instance, the Decision Transformer (DT) \citep{chen2021decision} employs an autoregressive approach to predict actions conditioned on desired outcomes and historical states and actions, treating trajectories as sequences. Similarly, the Trajectory Transformer \citep{janner2021offline} demonstrates the ability of transformers to model trajectories for single-task policies. While these methods achieve strong results on pre-collected datasets, they are largely limited to reproducing expert behavior, lacking mechanisms for generalization to novel tasks or for improving sub-optimal datasets. Extensions like the Multi-Game Decision Transformer (MGDT) \citep{lee2022multi} and Gato \citep{reed2022generalist} generalize DT’s paradigm to multi-task and cross-domain settings. However, MGDT requires task-specific fine-tuning, while Gato’s reliance on expert demonstrations limits its applicability to environments with rich expert data.

Bayesian methods offer a principled framework for incorporating uncertainty and prior knowledge into RL \citep{ghavamzadeh2015bayesian}. The Bayes-Adaptive Markov Decision Process (BAMDP) framework \citep{martin1967bayesian, duff2002optimal} theoretically provides an efficient mechanism by maintaining a belief distribution over environment dynamics and rewards. However, the computational intractability of exact BAMDP solutions restricts their applicability to simple tasks \citep{asmuth2012learning, guez2012efficient, guez2013scalable, brunskill2012bayes, poupart2006analytic}. Recent work has proposed approximate methods to make Bayesian RL more tractable, such as VariBAD \citep{varibad}, RoML \citep{greenberg2024train}, and BOReL \citep{dorfman2021offline}. For instance, VariBAD leverages a variational autoencoder (VAE) to infer a belief distribution over tasks and dynamically adapts policies during online interaction, showing improvements in task generalization.

Meta-RL methods focus on enabling agents to ``learning to learn," commonly referred to as meta-learning, aiming to adapt to new tasks efficiently based on prior experience. Specifically, recent approaches to ICRL can be classified as in-context meta-RL techniques. In-context meta-RL approaches, such as Algorithm Distillation (AD) \citep{laskin2022context}, demonstrate the ability to improve performance iteratively by using multiple historical trajectories. Decision-Pretrained Transformer (DPT) \citep{lee2024supervised} extends this idea by learning reward and transition dynamics from unsorted trajectories and performing posterior sampling under specific conditions. More recently, Goal-Focused Transformer (GFT) \citep{dippel2024contextual} addresses task-specific generalization by incorporating goal information post hoc into context prompts, improving performance in environments with well-defined goals. However, GFT depends on extensive meta-knowledge for prompt construction, limiting its applicability across diverse environments.

Our proposed approach, \ourname (\acronym), draws inspiration from these prior works by combining belief modeling with the ICL capabilities of transformers. Unlike DPT, \acronym leverages a VAE to infer a belief distribution over reward functions. This belief is used as an explicit prompt to the transformer policy, providing a structured way to handle reward uncertainty. Compared to GFT, which incorporates task information implicitly post hoc, \acronym integrates the belief distribution directly into the transformer policy prompt, avoiding reliance on meta-knowledge and enhancing generalization across diverse environments. Furthermore, while VariBAD adapts dynamically through online interaction, \acronym operates in an offline ICL setting, using a pre-trained belief distribution to guide decision-making during online evaluation. This eliminates the need for additional online exploration during training, making it more suitable for environments where online data collection is impractical. By combining the scalability of transformer-based ICL with the robustness of belief augmentation, \acronym achieves superior performance in both online adaptation and task generalization, surpassing existing baselines such as DPT and GFT in the tested benchmark environments.

\section{Experiments}

In this section, we elaborate on the characteristics of \ourname, demonstrating its online performance in three representative and widely studied settings: 1) a simplified grid-world domain with a discrete action space, 2) a 3D picture-based maze environment, and 3) a continuous-control benchmark with high-dimensional state and action spaces modeling physics-based locomotion task. These environments, which are commonly used in prior works, serve as benchmarks to evaluate the adaptability and effectiveness of online reinforcement learning methods. 
For each setting, we discuss baselines, training, and evaluation protocols. Our findings reveal that \ourname\ can adapt to tasks during the initial rollout and collect, on average, higher returns compared to its baseline competitors. 

\subsubsection{Environments} 

To analyze the capabilities of the Heuristic Transformer (HT), we start with the Darkroom environment \citep{chevalier2018minimalistic, jain2020generalization}, where the agent must find a hidden goal in a sparse-reward gridworld. We also test a larger version, Darkroom Hard, to evaluate scalability and a stochastic variant of darkroom. Next, we use Miniworld \citep{chevalier2018miniworld}, a 3D visual navigation task with image-based observations and hidden objectives. To assess performance under more complex dynamics, we include four standard continuous control tasks from MuJoCo \cite{todorov2012mujoco}: Hopper, Walker2d, HalfCheetah, and Swimmer. Full environment details and dataset specifications are provided in the Appendices.

\subsubsection{Baselines}

To evaluate the in-context learning capabilities of \ourname\ (HT), we compare it against several relevant baselines. These include PPO \cite{schulman2017proximal} and SAC \cite{haarnoja2018soft}, which serve as the underlying expert policies used to generate offline datasets in MuJoCo. We also include $RL^{2}$ \citep{duan2016rl}, an online meta-RL method that optimizes adaptation across episodes, although it benefits from online interaction during training—unlike \acronym, which is trained purely offline. As such, we treat $RL^{2}$'s performance as a soft upper bound.

The Decision-Pretrained Transformer (DPT) \citep{lee2024supervised} and the Goal-Focused Transformer (GFT) \citep{dippel2024contextual} are closely related ICRL methods. DPT relies solely on pre-training from unsorted offline data, while GFT augments its prompt with goal-specific cues using meta-knowledge. \acronym improves upon both by inferring a belief over rewards via a separate VAE, enabling stronger generalization without requiring privileged information.

\subsubsection{Evaluation}

We evaluate our method, HT, against baseline approaches (DPT, GFT, and $RL^2$) across a diverse set of environments, including Darkroom (both the standard and more challenging Hard variant), stochastic versions of Darkroom, as well as Miniworld and MuJoCo tasks. Our evaluation focuses on online performance during test phases and the impact of varying amounts of pre-training. Results are reported as the mean return across multiple trials, with error bars indicating the $\pm$ standard deviation. To test the performance of \acronym, a new task $\tau$ is sampled from the test-task distribution $T_\text{{test}}$. The posterior belief $q(m|\eta_{:h})$ is conditioned on the in-context dataset $D_{\text{pre}}$ and subsequently used by policy model $M_{\theta}$. The in-context policy learned by the model is represented conditionally as $M_{\theta}(\cdot|\cdot, \cdot, D)$. Specifically, the policy is evaluated by selecting an action $a$, which is subsequently appended to $D$. This process is repeated over successive episodes until the specified number of episodes is completed. A notable feature of the testing phase is the absence of parameter updates to $M_{\theta}$ and $q_{\phi}$. The model $M_{\theta}$ performs a forward pass to compute a distribution over actions, conditioned on the in-context dataset $D$, the current belief $b_{h}$, and the query state $s_{h}$. The optimal action $a^{\star}$ is selected as $a_{h} \in arg max_{a}M_{\theta}(a|D, b_{h}, s_{h})$ when the learner visits state $s_{h}$. For online testing across multiple episodes of interactions, the dataset $D$ is initialized as empty, i.e., $D=\{\}$. At each episode, the model $M_{\theta}(\cdot|\cdot, \cdot, D)$ is deployed, where it samples actions $a_{h} \sim M_{\theta}(D, b_{h}, s_h)$ after observing the query state $s_h$. Throughout the episode, the model gathers interactions in the form $\eta_{:H} = \{(s, a, r, s')_{1}, \dots, (s, a, r, s')_{H}\}$. Once the maximum capacity of $D$ has been reached, e.g., $H=100$, the oldest interactions will be replaced with the latest.


\begin{figure}[t!]
  \centering

  \begin{subfigure}{0.45\linewidth}
    \centering
    \includegraphics[width=\linewidth]{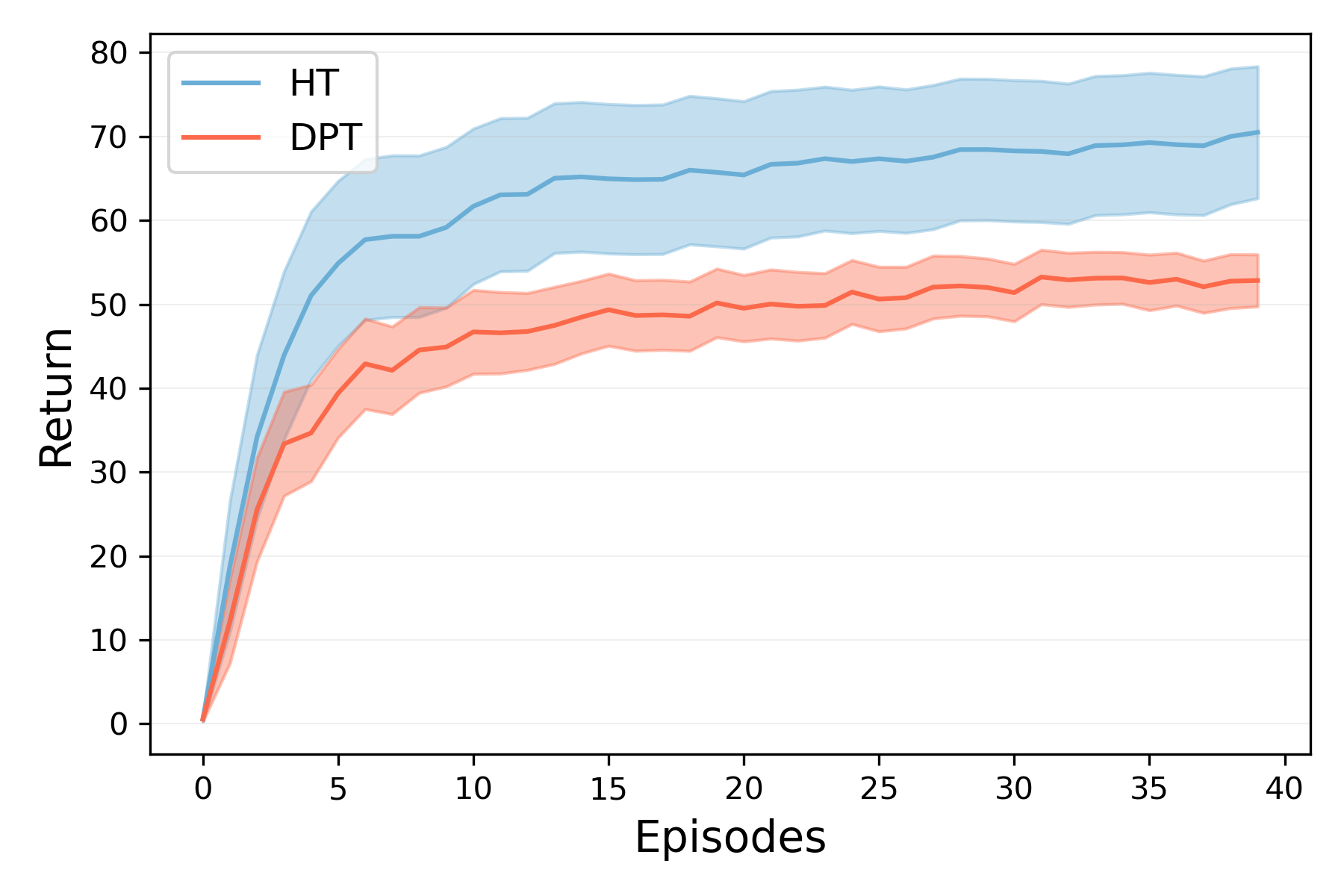}
    \caption{20\%}
    \label{fig:stochastic:20}
  \end{subfigure}\hfill
  \begin{subfigure}{0.45\linewidth}
    \centering
    \includegraphics[width=\linewidth]{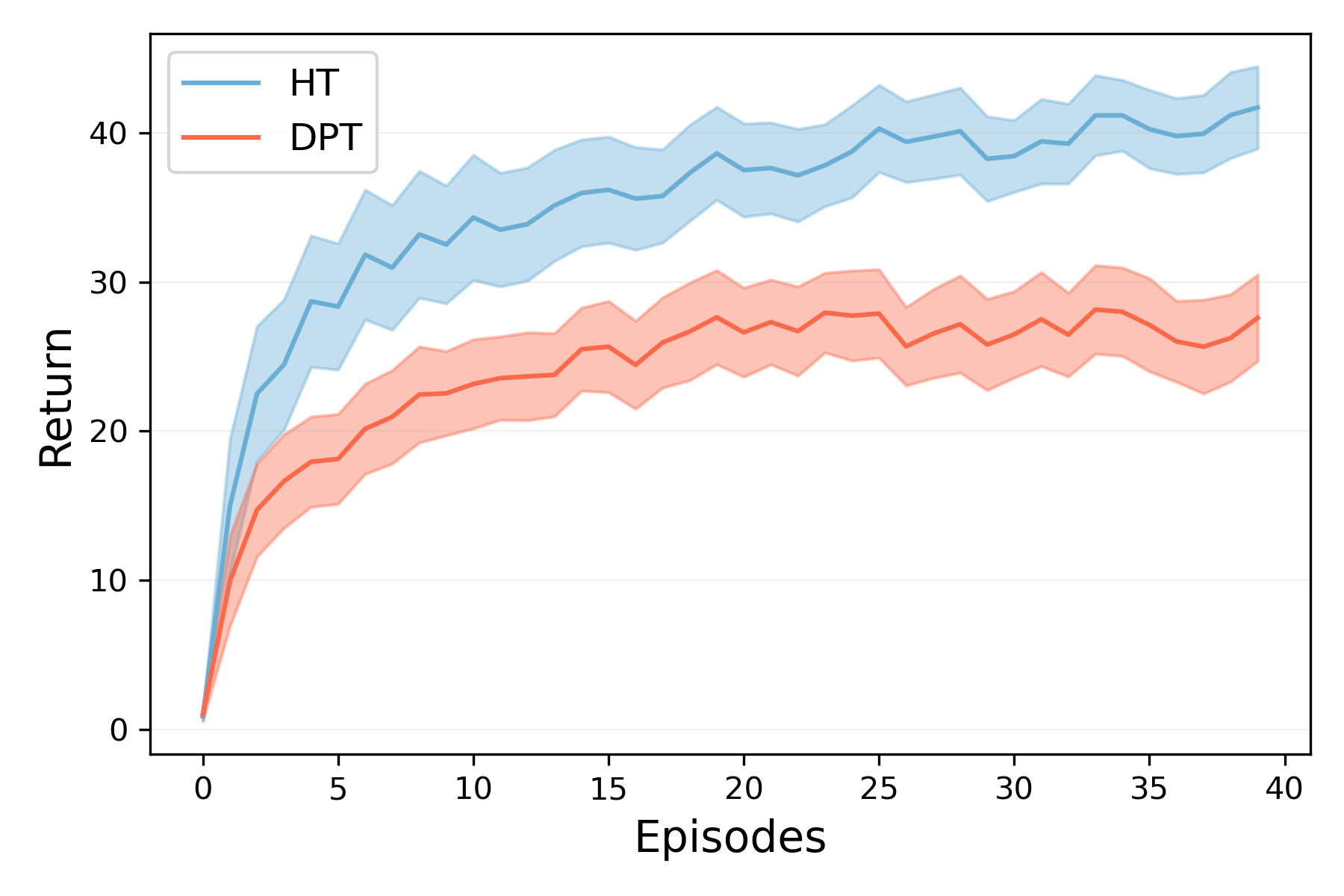}
    \caption{40\%}
    \label{fig:stochastic:40}
  \end{subfigure}

  \caption{Online performance on test goals in the Darkroom Stochastic environment under varying levels of transition noise. 
  (a) 20\% random action misdirection. 
  (b) 40\% random action misdirection. 
  Results are mean return $\pm$ std, averaged over 20 trials and 5 seeds.}
  \label{fig:stochastic}
\end{figure}

\subsubsection{Darkroom} 

In the standard Darkroom setting, \acronym demonstrates promising performance, as shown in Figure \ref{fig:darkroom}(a). During online evaluation, \acronym achieves faster learning and higher mean returns compared to the baselines, particularly in the earlier episodes. These results suggest that \acronym effectively adapts to new test goals in this environment. Figure \ref{fig:darkroom}(b) highlights the effect of pre-training on Darkroom performance. Across 100, 200, 500, and 1000 training epochs, \acronym consistently achieves higher returns than DPT and GFT, indicating a faster adaptation towards generalization to unseen tasks. In the more challenging Darkroom Hard, \acronym continues to show competitive performance compared to the baselines. Note that the context size for Darkroom Hard quadrupled, while the environmental complexity increased by a factor of 16. Figure \ref{fig:darkroom}(c) illustrates that during online evaluation, \acronym steadily improves and achieves significantly better returns than GFT and DPT, both of which struggle to progress over time or even find the goal location. As shown in Figure \ref{fig:darkroom}(d), \acronym consistently improves as the number of pre-training epochs increases, while DPT hardly benefits from additional training. This also shows the difficulty of solving Darkroom Hard. While DPT shows good performance in Darkroom, it fails to solve Darkroom Hard under the given conditions. These results suggest that \acronym is capable of handling increased environmental complexity better than GFT and DPT.

\subsubsection{Transition Uncertainty}

To assess the robustness of \acronym in stochastic environments, we introduce a variant of the Darkroom environment with controlled transition noise. In this setting, we introduce a stochastic perturbation to the agent’s actions. Specifically, when the agent intends to move in a cardinal direction (e.g., up), there is a fixed probability—either 20\% or 40\%—that the agent instead moves left or right with equal likelihood. This modification does not alter the reward structure but increases the difficulty of reliably reaching the goal, thereby testing the model’s ability to infer and adapt to noisy dynamics. Such controlled randomness enables us to systematically evaluate the generalization and in-context adaptation capabilities of \acronym in the presence of stochastic transitions. Figure~\ref{fig:stochastic}(a) shows that \acronym experiences only a minor drop in performance under 20\% action misdirection, performing nearly on par with the vanilla Darkroom environment. Even with a higher level of stochasticity—where nearly half of the actions are misdirected, as shown in Figure~\ref{fig:stochastic}(b)—\acronym still maintains reasonable performance, while continuing to outperform DPT.


\begin{figure}[t!]
  \centering

  \begin{subfigure}{0.45\linewidth}
    \centering
    \includegraphics[width=\linewidth]{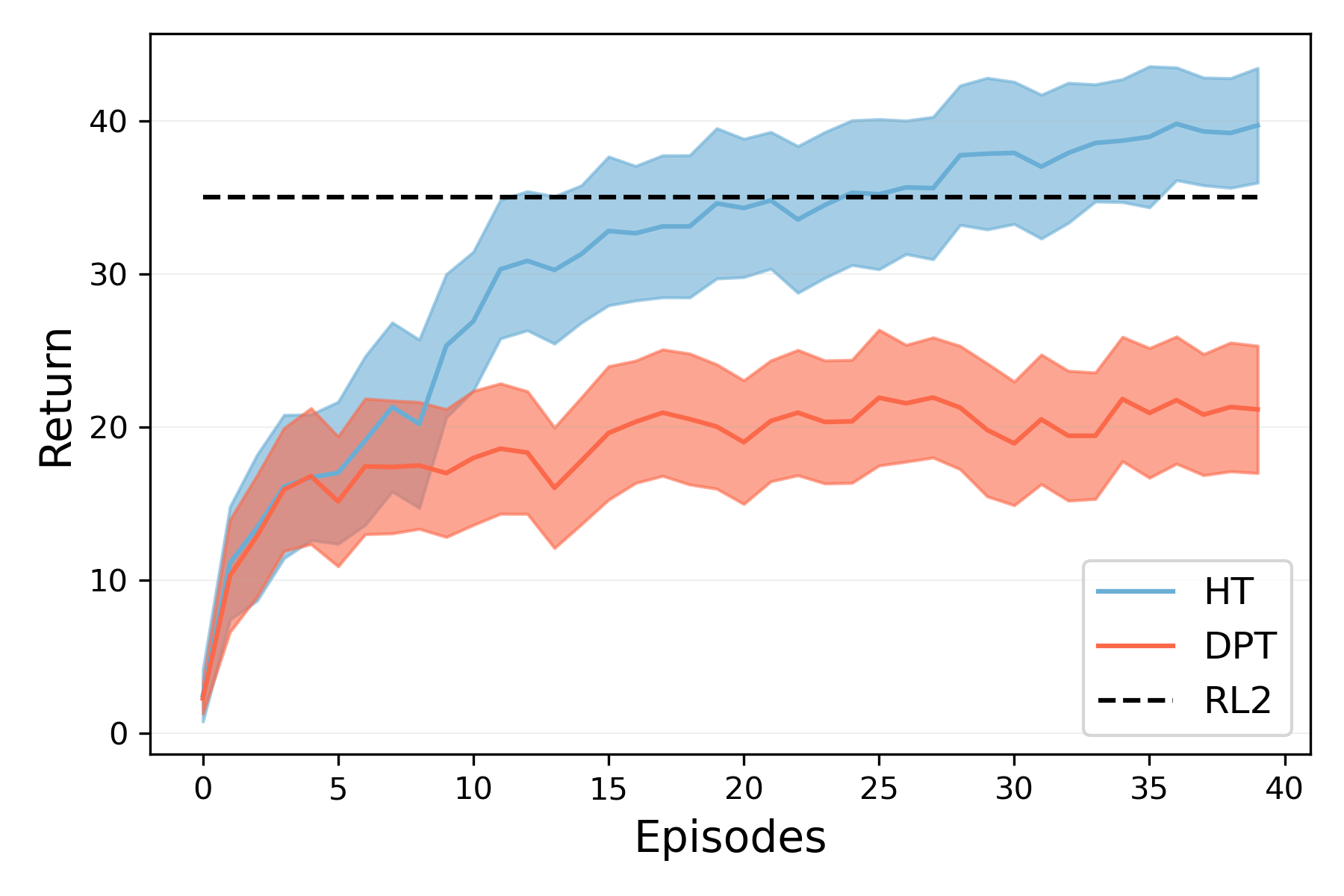}
    \caption{Miniworld}
    \label{fig:miniworld:a}
  \end{subfigure}\hfill
  \begin{subfigure}{0.45\linewidth}
    \centering
    \includegraphics[width=\linewidth]{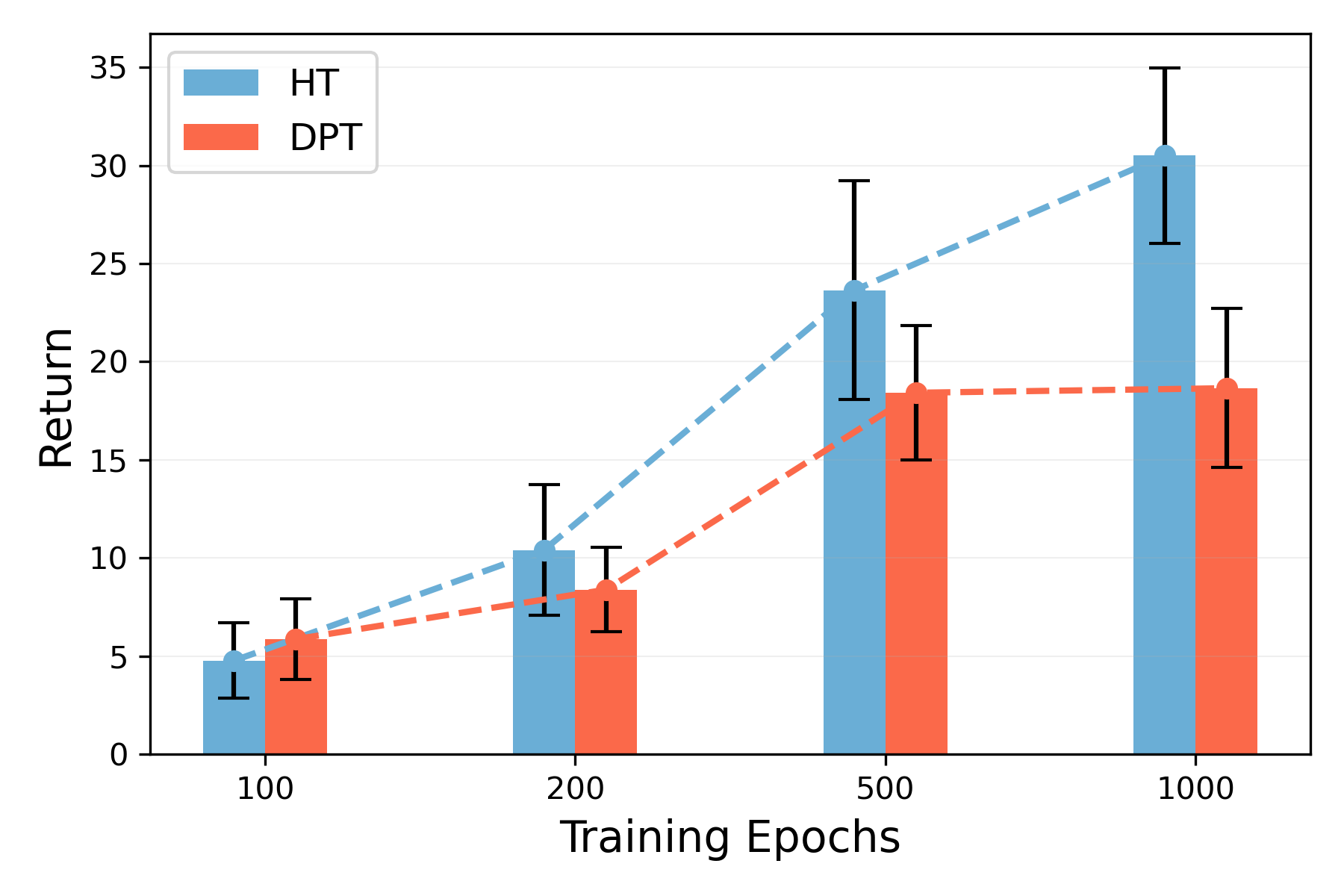}
    \caption{Miniworld (epochs)}
    \label{fig:miniworld:b}
  \end{subfigure}

  \caption{(a) Online performance on test goals in Miniworld. 
  (b) Online performance on test goals in Miniworld, after certain pre-training epochs. 
  Results are the mean return $\pm$ standard deviation on 20 trials across 5 seeds.}
  \label{fig:miniworld}
\end{figure}

\subsubsection{Miniworld}

We then evaluate \acronym and baseline methods in the Miniworld environment, which aims to demonstrate \acronym's capability to perform in image-based environments. We excluded GFT as a baseline comparison for the Miniworld environment, as GFT is designed for discrete state spaces, whereas Miniworld has a continuous state space. Figure \ref{fig:miniworld}(a) shows that \acronym achieves significantly faster adaptation and higher returns during online evaluation compared to DPT. While both methods improve over time, \acronym appears to reach higher performance levels more consistently.
Figure \ref{fig:miniworld}(b) shows the effect of pre-training amount on algorithm performance in Miniworld. Across all pre-training epochs, \acronym performs competitively, achieving higher returns than DPT. These results indicate that \acronym can effectively leverage pre-training to improve adaptation and generalization in this environment.

\subsubsection{MuJoCo}

Finally, we evaluate \acronym's performance in four MuJoCo environments from OpenAI Gym \cite{DBLP:journals/corr/BrockmanCPSSTZ16}. While \acronym is explicitly designed to generalize across varying MDPs, we include MuJoCo to test whether its in-context learning capabilities remain effective in high-dimensional continuous control tasks, even when task variation is limited. In this setup, a key challenge is that each environment corresponds to a fixed MDP—meaning the training datasets do not exhibit variation in reward or transition functions. As a result, there is no explicit task uncertainty to resolve, and the potential benefit of modeling a reward belief distribution may be reduced. In such cases, \acronym could rely more on memorization of environment-specific patterns than true generalization. Nonetheless, Table~\ref{tab:mujoco_results} shows that \acronym performs competitively across all environments, indicating that its architecture can still leverage contextual information effectively even in the absence of explicit MDP shifts.
The baseline methods (PPO and SAC) are the original algorithms used to generate the offline training datasets for \acronym and DPT. HT-P and HT-S represent the \ourname trained solely on PPO and SAC rollouts, respectively, while HT-SP uses a 50/50 mixture of both. The same setup applies to DPT-SP. Bold values indicate the best-performing baseline and ICRL method per environment. The results demonstrate that HT-SP consistently achieves strong performance across environments, outperforming DPT-SP. Our findings suggest that \acronym generally benefits from greater diversity in the training data, as seen in the improved performance of the HT-SP variant across most environments. We slightly adapted the VAE architecture to better accommodate the characteristics of the MuJoCo environments. In particular, these tasks involve larger state spaces and require longer context sizes (H = 2000). A detailed description of the architectural changes, together with the full set of hyperparameters and implementation specifics, is provided in the appendix.

Our experimental results demonstrate \acronym's superior performance over baselines in Darkroom, Miniworld and MuJoCo particularly in terms of online adaptation and generalization after pre-training. While the baseline methods in Darkroom and Miniworld also demonstrate improvement in certain cases, \acronym consistently achieves higher returns across a range of task settings.

\begin{table}[t]
\scriptsize
\centering
\begin{tabular}{@{}lcccc@{}}
\toprule
\textbf{Algorithm} & \textbf{Hopper} & \textbf{Walker2d} & \textbf{HalfCheetah} & \textbf{Swimmer} \\
\midrule
PPO     & 1710.8 ± 523.9  & 2267.6 ± 1020.8 & 1646.7 ± 108.1 & 119.1 ± 2.2 \\
SAC     & \textbf{1839.3 ± 164.8}  & \textbf{5252.4 ± 51.5}  & \textbf{2328.1 ± 11.9} & \textbf{143.5 ± 4.9} \\
HT-P    & 1494.4 ± 431.9  & 1900.4 ± 890.6  & 1524.1 ± 124.7 & 115.2 ± 3.9 \\
HT-S    & 1541.3 ± 384.2  & 3020.5 ± 408.8  & 1905.4 ± 56.7  & 121.9 ± 3.7 \\
HT-SP   & \textbf{1711.5 ± 317.1}  & \textbf{3565.2 ± 433.2}  & \textbf{1968.3 ± 60.5}  & \textbf{133.0 ± 3.9} \\
DPT-SP  & 1620.1 ± 313.9  & 3099.3 ± 432.7  & 1878.8 ± 61.0  & 123.5 ± 5.3 \\
\bottomrule
\end{tabular}
\caption{Evaluation (Mean ± Std) over 100 episodes on MuJoCo environments across 5 seeds}
\label{tab:mujoco_results}
\end{table}

\subsection{Further Experimental \& Ablation Results}

\begin{figure}[h!]
  \centering

    \includegraphics[width=\linewidth]{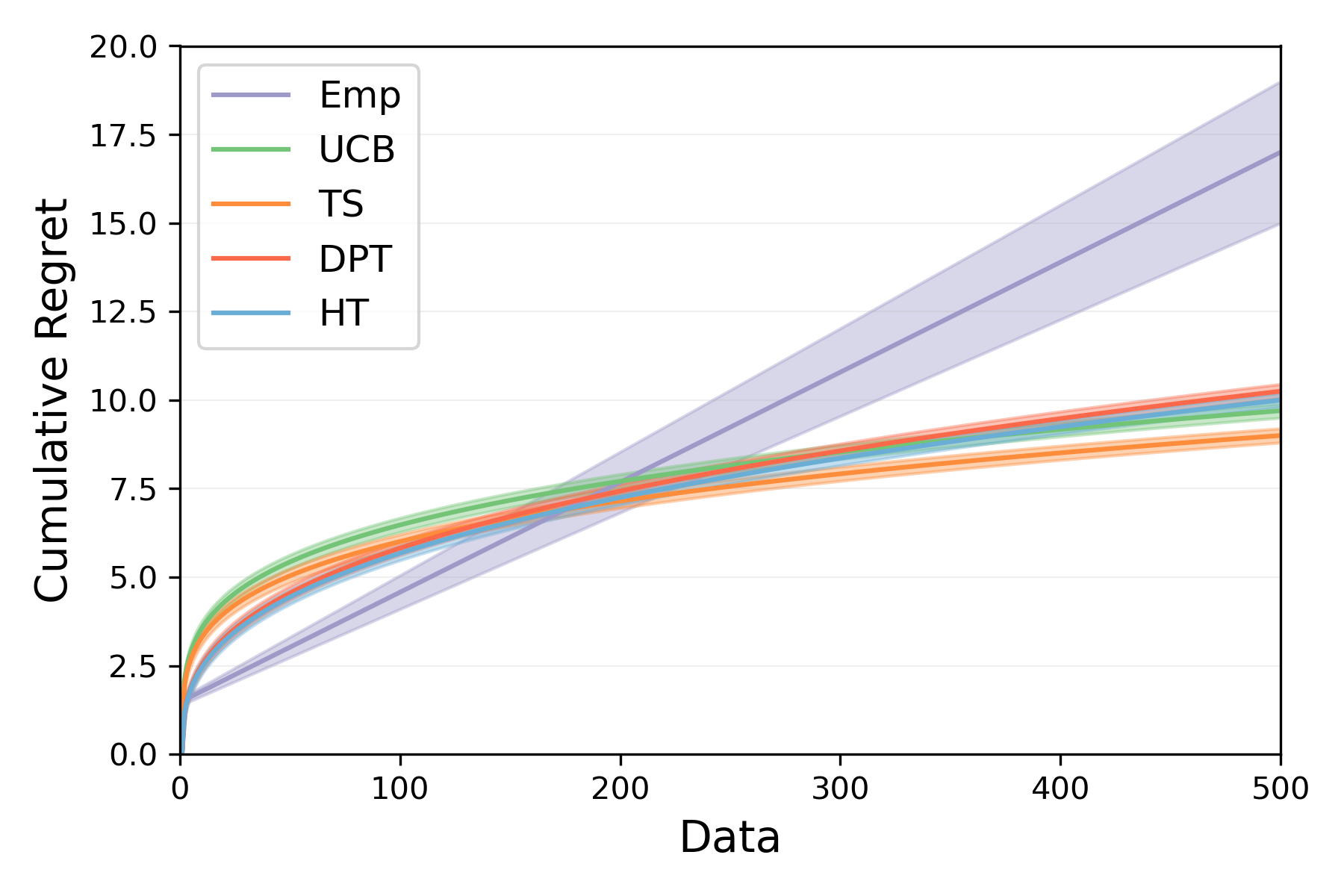}
 
  \caption{ Online cumulative regret in the Bandit environment. 
  Reported are the mean return $\pm$ standard deviation on 20 trials across 10 seeds.}
  \label{fig:ablation_1}
\end{figure}

We evaluate \acronym in the Bandit setting to test whether it can learn useful decision-making priors in the absence of temporal structure or long-term dependencies. In Bandit the agent is tasked with selecting one of five arms, each associated with an unknown reward distribution. The means of these distributions are randomly sampled from a uniform distribution, and the agent must infer the optimal arm based on a limited in-context dataset containing action-reward pairs. Bandit tasks offer minimal context—single-step episodes with independent rewards per arm — we set $H = 1$. This setting is deliberately challenging for a belief-based model like \acronym, as the short horizon limits the value of modeling uncertainty over reward distributions. Indeed, we do not expect HT to outperform specialized baselines here. Nevertheless, as shown in Figure~\ref{fig:ablation_1}, \acronym achieves comparable performance to classical exploration-based methods like UCB and Thompson Sampling (TS), while slightly outperforming DPT. Emp selects the arm with the highest empirical mean, and UCB favors underexplored arms with pessimistic estimates. In contrast, UCB and TS are designed for efficient exploration, which explains their strong performance. That \acronym matches these methods despite not being explicitly trained for exploration suggests it has learned a generalizable decision-making heuristic that remains effective even in simple, highly-optimized settings.


\subsubsection{Ablation Results}

\begin{figure}[h!]
  \centering
  \begin{subfigure}{0.45\linewidth}
    \centering
    \includegraphics[width=\linewidth]{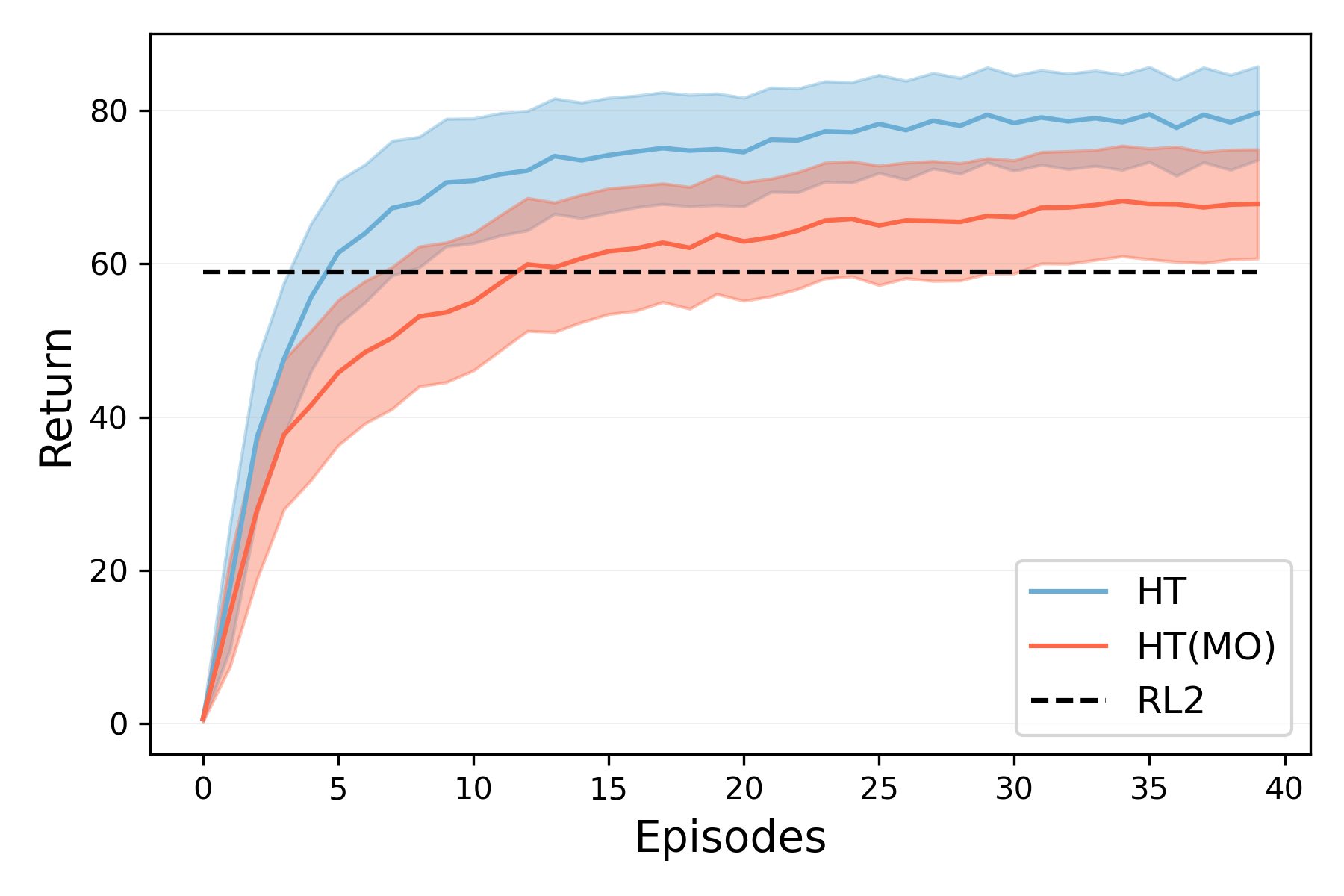}
    \caption{Darkroom}
    \label{fig:ablation:darkroom}
  \end{subfigure}\hfill
  \begin{subfigure}{0.45\linewidth}
    \centering
    \includegraphics[width=\linewidth]{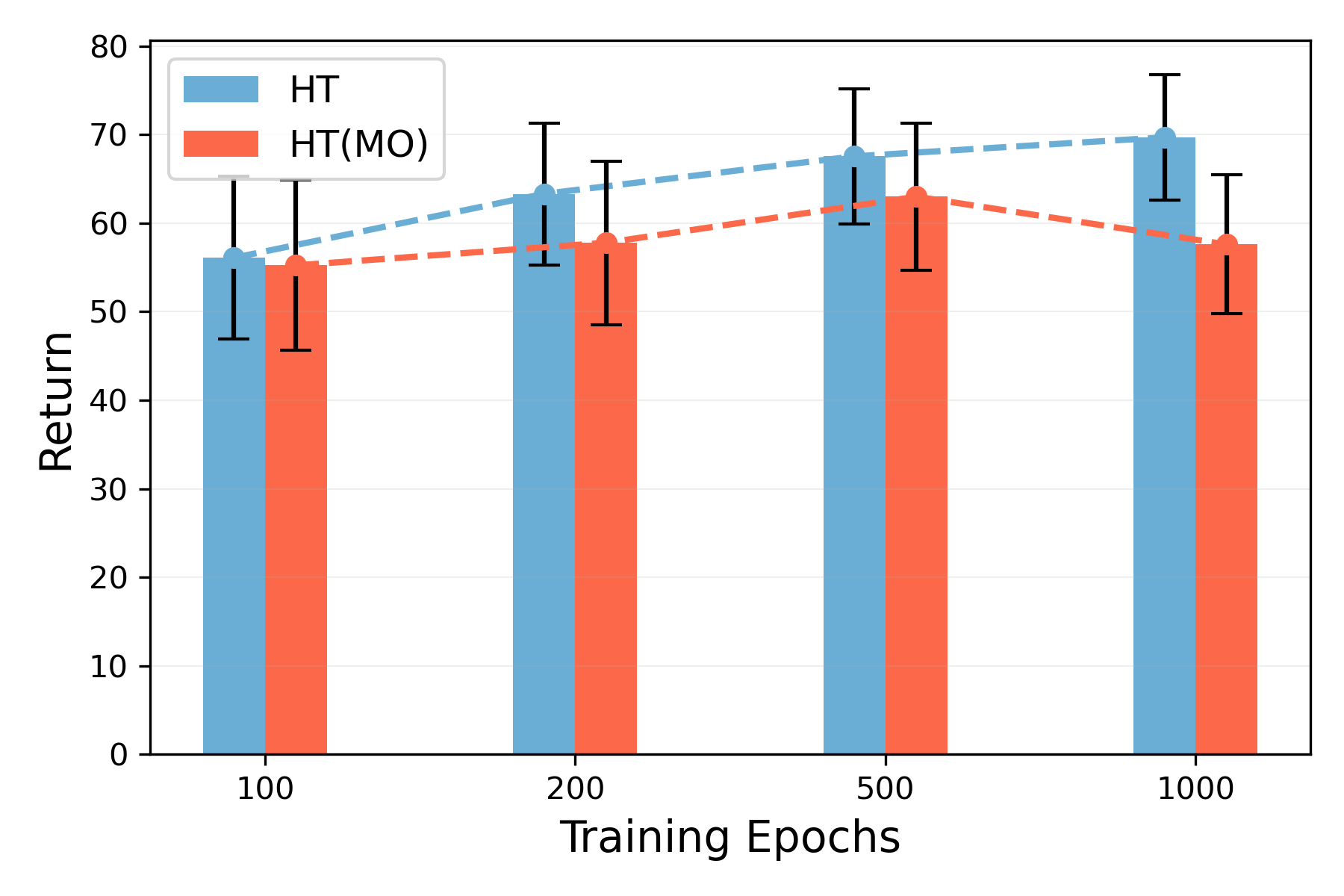}
    \caption{Darkroom (epochs)}
    \label{fig:ablation:darkroom-epoch}
  \end{subfigure}

  \caption{
  (a) Online performance on test goals in Darkroom between HT and HT (MO). 
  (b) Online performance on test goals in Darkroom, after certain pre-training epochs between HT and HT (MO). 
  Reported are the mean return $\pm$ standard deviation on 20 trials across 10 seeds.}
  \label{fig:ablation_2}
\end{figure}

Incorporating a multi-objective loss function within the transformer framework is both theoretically sound and practically effective for modeling the belief distribution $b$. This approach leverages the transformer's ability to capture complex dependencies, enabling the seamless integration of Bayesian reasoning within the policy model $M_{\theta}$, thereby rendering the first pre-training phase redundant. By incorporating a multi-objective loss, the model can simultaneously maintain a belief distribution over rewards and refine its policy, effectively closing the loop between environment inference and optimal decision-making. This synergy enhances the model's capacity to generalize across tasks by capturing reward dynamics more effectively. As a result, this framework represents a principled and computationally efficient method for approximating what we term the Heuristic Bayesian Policy. In such a framework the loss function would be:

\begin{equation}\label{eq:ablation}
\begin{split}
    \mathcal{L}_{3}(\theta, \theta_{\text{action}},\theta_{\text{reward}}) = \\ \min_{\theta} \mathbb{E}_{P_{\text{pre}}} \sum_{j \in [n]} \Big[-\log\big( \lambda_{1} M_{\theta_{\text{action}}, \theta}(a^\ast | D_j, s_{\text{query}}) + \\ \lambda_{2} M_{\theta_{\text{reward}}, \theta}(r | D_j)\big) \Big],
\end{split}
\end{equation}

where $M_{\theta_{\text{action}}, \theta}$ and $M_{\theta_{\text{reward}}, \theta}$ are the action and reward prediction heads with their respective parameters and the shared parameters $\theta$, with $\lambda_{1}, \lambda_{2}$ being weighting parameters. We refer to this model as \ourname(MO). Figures \ref{fig:ablation_2}(a) and (b) show the mean episode return achieved by \ourname and \ourname(MO) on 20 trials across 10 seeds in Darkroom. While \ourname(MO) is theoretically plausible, empirical results do not show it performing on par with \ourname. Additional implementation details and hyperparameters are provided in the appendix.

\section{Conclusion \& Future Work}

In this work, we introduced \ourname, an in-context reinforcement learning (ICRL) method that augments transformer policies with a learned belief distribution over rewards. By conditioning on a VAE-inferred belief, a query state, and an in-context dataset, \ourname improves online adaptation and generalization over baseline methods in various domains.

Our experiments across Darkroom, Miniworld and MuJoCo show that \ourname achieves strong performance, particularly in environments with high complexity or sparse rewards. It adapts quickly, scales to visual tasks, and maintains robustness under stochastic transitions.

However, \ourname depends on high-quality pre-training data and currently requires access to optimal actions during training. Relaxing this assumption, as explored in \citep{lee2024supervised}, remains an important direction. Additionally, future work could explore generalization across environments—for instance, by training on multiple MuJoCo domains to assess whether \ourname can capture shared structure and transfer effectively across different MuJoCo tasks.

Overall, our findings suggest that integrating belief distributions into transformer-based ICRL offers a promising path forward for learning adaptive, generalizable policies.


\bibliographystyle{plainnat}
\bibliography{main} 

\appendix
\onecolumn
\section{Appendices}

\subsection{Additional Related Work}

The idea of In-context meta-RL techniques, to enable agents to self-improve has been explored in RL for decades, although early efforts were primarily focused on optimizing hyperparameters \citep{ishii2002control}. In-context meta-RL methods \citep{wang2016learning, duan2016rl} are typically developed within an online framework, leveraging memory-based architectures to maximize multi-episodic value functions through direct interactions with the environment. Another branch of online meta-RL focuses on identifying optimal initial network parameters, which are then refined through additional gradient updates \citep{finn2017model, nichol2018first}. Recent advancements in meta-RL have delivered significant progress, including improved performance on established benchmarks and extensions into offline domains such as Bayesian RL \citep{dorfman2021offline} and optimization-driven meta-RL \citep{mitchell2021offline}. Given the challenges associated with fully offline learning, hybrid offline-online approaches have emerged as a promising direction \citep{zahavy2020self, pong2022offline}.

Contextual Markov Decision Processes (CMDPs) can be understood as predecessors to in-context RL, where the task description is treated as a context on which environment dynamics and rewards depend \citep{hallak2015contextual, jiang2017contextual, dann2019policy, modi2019contextual}. ICRL addresses tasks by leveraging prompts or demonstrations to guide decision-making \citep{chen2021decision, janner2021offline}. Transformer-based policies, when trained at scale, often exhibit in-context learning capabilities \citep{lee2022multi, reed2022generalist}. This approach operates entirely within the context of the provided data, without requiring updates to the neural network parameters. Pioneer work from the aforementioned \citep{chen2021decision} gave inspiration from which numerous variants have emerged e.g.
\citep{yamagata2023q, zisman2023emergence, sinii2023context, huang2024context, shi2023cross, ghanem2023multi, liu2023emergent, huang2024decision}.

\subsection{\emph{ELBO} Derivation}

Equation \ref{eq:2} can be derived as follows: 

\begin{align}
    \mathbb{E}_{\rho_{\tau}(D)} [\log p_\Phi(r_{:h})] 
    &= \mathbb{E}_{\rho_{\tau}} \left[ \log p_\Phi(r_{:h}) \right] \notag \\
    &= \mathbb{E}_{\rho_{\tau}} \left[ \log \int p_\Phi(r_{:h}, m) \, dm \right] \notag \\
    &= \mathbb{E}_{\rho_{\tau}} \left[ \log \int q_\phi(m \mid \eta_{:h}) \frac{p_\Phi(r_{:h}, m)}{q_\phi(m \mid \eta_{:h})} \, dm \right].
\end{align}

Using Jensen's inequality:
\begin{align}
    \mathbb{E}_{\rho_{\tau}(D)} [\log p_\Phi(r_{:h})] 
    &\geq \mathbb{E}_{\rho_{\tau}} \left[ \int q_\phi(m \mid \eta_{:h}) \log \frac{p_\Phi(r_{:h}, m)}{q_\phi(m \mid \eta_{:h})} \, dm \right] \notag \\
    &= \mathbb{E}_{\rho_{\tau}} \left[ \mathbb{E}_{q_\phi(m \mid \eta_{:h})} \left[ \log p_\Phi(r_{:h}, m) - \log q_\phi(m \mid \eta_{:h}) \right] \right].
\end{align}

Decomposing the joint probability \( \log p_\Phi(r_{:h}, m) \):
\begin{align}
    \log p_\Phi(r_{:h}, m) &= \log p_\Phi(r_{:h} \mid m) + \log p_\Phi(m).
\end{align}

Substituting this back:
\begin{align}
    \mathbb{E}_{\rho_{\tau}(D)} [\log p_\Phi(r_{:h})] 
    &\geq \mathbb{E}_{\rho_{\tau}} \left[ \mathbb{E}_{q_\phi(m \mid \eta_{:h})} \left[ \log p_\Phi(r_{:h} \mid m) + \log p_\Phi(m) - \log q_\phi(m \mid \eta_{:h}) \right] \right] \notag \\
    &= \mathbb{E}_{\rho_{\tau}} \left[ \mathbb{E}_{q_\phi(m \mid \eta_{:h})} \left[ \log p_\Phi(r_{:h} \mid m) \right] - \text{KL}(q_\phi(m \mid \eta_{:h}) \| p_\Phi(m)) \right].
\end{align}

Defining the Evidence Lower Bound (ELBO):
\begin{align}
    \text{ELBO}_{h} &= \mathbb{E}_{\rho_{\tau}} \left[ \mathbb{E}_{q_\phi(m \mid \eta_{:h})} [\log p_\Phi(r_{:h} \mid m)] - \text{KL}(q_\phi(m \mid \eta_{:h}) \| p_\Phi(m)) \right].
\end{align}

Thus, the final inequality becomes:
\begin{align}
    \mathbb{E}_{\rho_{\tau}(D)} [\log p_\Phi(r_{:h})] \geq \text{ELBO}_{h}.
\end{align}

\subsection{Model Architecture Description}

In this section, we present a detailed explanation of the architecture introduced in Section \ref{sec:mod} and illustrated in Figure \ref{fig:model}. Hyperparameter specifics for the models are discussed within their respective sections. The implementation is carried out in Python using PyTorch \citep{paszke2019pytorch}. To construct the belief we use a Variational Autoencoder, the backbone of the transformer policy model is an GPT-2 architecture, sourced from the HuggingFace Transformers library.

For clarity, let us assume that $S$ and $A$ are subsets of $\mathbb{R}^{d_{S}}$ and $\mathbb{R}^{d_{A}}$, respectively. Discrete state and action spaces are represented through one-hot encoding. Consider a single training example derived from a (potentially unknown) task $\tau$. The available data consists of a dataset $D$ of interactions within $\tau$, a query state $s_{\text{query}}$, and its corresponding optimal action $a^{\star} = \pi^{\star}_{\tau}(s_{\text{query}})$.

\subsubsection{Variational Autoencoder}

To prepare the input embeddings for the Variational Autoencoder architecture, we process the dataset $D=\{(s_{j}, a_{j},  r_{j}, s'_{j} )\}$. For each transition tuple, we construct a sequence by stacking the context information $(s_{j}, a_{j}, r_{j}, s'_{j})$ along the last dimension, forming a tensor of shape $[, d_{input}]$, where $d_{input}:=2d_{S} + d_{A}+1$. These stacked sequences serve as input to the encoder network.

The encoder processes the flattened input through multiple layers to produce latent representations. Specifically, the encoder maps the input into a latent space, characterized by a mean vector $\mu_{j}$ and a log-variance vector $\log \sigma_{j}^2$, which define a Gaussian distribution over the latent space. To obtain the latent representation $m_{j}$ for each transition, we apply the reparameterization trick, sampling $m_{j}=\mu_{j} + \epsilon \times \sigma_{j}^2$, where $\epsilon \sim \mathcal{N}(0, I)$.

Once encoded, the latent variable $m$ is pooled across all transitions to form a single global latent code per batch, ensuring the model captures contextual task information across all transitions in $D$. This pooled $m$ is then concatenated with each transition tuple $(s_{j}, a_{j}, s'_{j})$ and passed through the decoder network to predict the corresponding reward $r_{j}$. The decoder operates on the concatenated vector $(b, s_{j}, a_{j}, s'_{j})$, mapping it back to the reward space.

The training loss is the sum of a reconstruction loss and a KL divergence term. The reconstruction loss depends on the assumed observation model: for environments with continuous rewards (e.g., MuJoCo and Bandit), we assume a \emph{Gaussian observation model} and use \emph{mean squared error (MSE)} between the predicted rewards $\hat{r}_{j}$ and the ground-truth rewards $r_{j}$. For environments with binary rewards (e.g., Darkroom and Miniworld), we assume a \emph{Bernoulli observation model} and use \emph{binary cross-entropy (BCE)} loss. The KL divergence regularizes the latent space by minimizing the divergence between the learned posterior $q(m \mid D)$ and a standard Gaussian prior $p(m)$, as defined in equation~\ref{eq:4}.

For image-based inputs, the dataset $D = \{(s_{j}, a_{j}, r_{j}, img_{j})\}_{j \in [n]}$ includes an additional component: $img_{j}$, which represents image observations associated with the transitions. Each image is processed through a convolutional neural network (CNN) to extract compact feature vectors. The extracted image features, along with $s_{j}, a_{j}, r_{j}$, are stacked into a tensor with dimensions $[n, d_{input}]$, where: $d_{input}:= d_{S} + d_{S} + 1 + d_{img}$. The CNN component, consisting of convolutional and fully connected layers, transforms raw image data into feature embeddings of dimension $d_{img}$. This embedding is concatenated with the non-visual features to form the complete input representation.

\subsubsection{Transformer Model}

To prepare the input embeddings for the GPT backbone, we process the dataset $D= \{(s_{j}, a_{j}, r_{j}, s'_{j}) \}_{j \in [n]}$. For each transition tuple, we construct vectors 
$\mathcal{V}_{j}=(s_{j}, a_{j}, r_{j}, s'_{j})$ by stacking the tuple elements into a vector of dimension $d_{\mathcal{V}}:= 2d_{S} + d_{A} + 1$. Additionally, we create a vector $v_{1}:=(s_{\text{query}}, 0)$ and $v_{2}:=(m)$, where $0$ is a zero vector of sufficient length to make its dimension equal to $\mathcal{V}$. These components form an input sequence $X=(v_{1}, v_{2},\mathcal{V}_{1}, \dots, \mathcal{V}_{n})$ of length $n+2$.

Since the dataset $D$ does not depend on a specific order, we omit positional encoding to respect this invariance. A linear layer $lin(X)$, is applied to the input sequence, and the resulting output is passed to the transformer. The transformer produces a sequence of outputs $Y=(\hat{y}_{0}, \hat{y}_{1}, \dots, \hat{y}_{n})$.

For continuous action spaces, the outputs $\hat{y}_{j}$ can be used directly to predict $a^{\star}$. For discrete action spaces, the outputs are treated as logits, which are converted into either action distributions over $A$ or one-hot vector predictions of $a^{\star}$. Specifically, we compute the action probabilities as $\hat{\varrho}_{j}=softmax(\hat{y}_j) \in A$, which represents the probability distribution over actions.

Due to the causal nature of the GPT architecture \citep{radford2019language}, the predicted probabilities $\hat{\varrho}_{j}$ depend only on $s_{\text{query}}$, the latent representation $m_{j}$, forming the posterior belief $b_{h}$ at point $h$ and the partial dataset $D_{j}=\{(s_{l}, a_{l}, s'_{l}, r_{l} )\}_{l \in [j]}$. This is why we express the model as $M_{\theta}(\cdot|D_{j}, b_{j}, s_{\text{query}}) = \hat{\varrho}(\cdot)$
to indicate that the prediction for the $j-th$ element relies on $D_{j}$, not the full dataset $D$ for the model $M$ with parameters $\theta \in \Theta$.

For $j=0$, the prediction of $a^{\star}$ is made using only the query state $s_{\text{query}}$ and the prior belief $b_{0}$, representing the prior over $a^{\star}$ without any additional contextual task information. The loss for this training example is calculated as the cross-entropy across all $j \in [l]$ is $-\sum_{j \in [l]} \log \hat{p}_{j}(a^{\star})$.

\subsection{Hyperparameter}

Throughout all transformer-based ICRL methods, we used consistent hyperparameter configurations to ensure better comparability. None of the methods underwent extensive hyperparameter tuning. For baselines, the hyperparameters were determined based on the original papers.

\subsubsection{DPT}. We used an open-sourced implementation from \citep{lee2024supervised} for DPT. The transformer architecture was configured with an embedding size of $32$, four hidden layers, and two attention heads per layer. The context length was set to $100$ steps for Darkroom and 400 for Darkroom Hard. In the Miniworld environment, input images were processed using a convolutional network with two convolutional layers (each with $16$ kernels of size $3\times3$ followed by a linear layer that produced an embedding of dimension eight. Beyond this preprocessing step, the transformer model retained the same hyperparameter settings as in the Darkroom environments. Optimization was performed using the AdamW optimizer with a weight decay of 1e-4, a learning rate of 1e-3, and a batch size of $64$.

\subsubsection{GFT} For GFT, we based our implementation on \citep{lee2024supervised} and adapted the architecture following \citep{dippel2024contextual}. The hyperparameters were kept consistent with those used for DPT, including an embedding size of 32, four hidden layers, two attention heads per layer, and a context length of 100 steps for Darkroom and 400 steps for Darkroom Hard. The AdamW optimizer with a weight decay of 1e-4, a learning rate of 
1e-3, and a batch size of $64$ was used.

\subsubsection{$RL^2$} The results for $RL^2$ were obtained using the open-source implementation from \citep{varibad}. This method employed Proximal Policy Optimization (PPO) \citep{schulman2017proximal} as the reinforcement learning algorithm and defined a trial as a sequence of four consecutive episodes. In the Darkroom environment, the policy architecture consisted of a single hidden layer with 32 units. For the Miniworld environment, the policy was parameterized using a convolutional neural network (CNN) with two convolutional layers (each containing $16$ filters of size $3\times3$, followed by a linear layer with an output dimension of eight.

\subsubsection{HT} The VAE model for HT was configured with a latent space of dimensionality 5 for Bandit, 10 for Darkroom and Darkroom Hard, and 15 for Miniworld. The encoder and decoder networks used two fully connected layers with sizes $64$ and $32$. In the Miniworld environment, input images were processed through a convolutional network with two convolutional layers (each with $16$ kernels of size $3\times3$) followed by a linear layer that produced an embedding of dimension eight. We used the Adam \citep{diederik2014adam} optimizer with learning rate  1e-3. The transformer model retained consistent hyperparameter settings, with an embedding size of $32$, four hidden layers, and two attention heads per layer. The context length was set to $1$ step for Bandit, $100$ steps for Darkroom and $400$ for Darkroom Hard. Optimization utilized the AdamW optimizer with a weight decay of 1e-4, a learning rate of 1e-3, and a batch size of $64$. We adapt the VAE architecture for MuJoCo environments by replacing the feedforward encoder with a bidirectional Gated Recurrent Unit (BiGRU) \cite{schuster1997bidirectional, cho2014learning}. The encoder processes sequences of length up to 2000 using a BiGRU with a hidden size of 32 per direction, yielding a 64-dimensional pooled representation. This representation is projected into a latent space of dimensions 26, 31, 31 and 19 respectively via fully connected layers that output the mean and log-variance of the posterior. The decoder receives the latent variable alongside each transition tuple $(s, a, s')$ and predicts the corresponding reward. The decoder uses two fully connected layers with sizes 64 and 32. We use the Adam optimizer with a learning rate of 1e-3.

\subsection{Training Datasets}\label{app:datasets}

Our study evaluates in both sparse-reward and high-dimensional settings. In the Darkroom environments, reward sparsity is inherent by design: in Darkroom, 36.28\% of training examples contain no reward signal within a context of 100 steps, and this increases to 87.65\% in Darkroom Hard (context length 400), posing a significant challenge for reward representation learning. To test generalization beyond low-dimensional MDPs, we include Miniworld—a 3D visual navigation task with RGB image inputs—demonstrating that our method scales to complex observation spaces. Additionally, we introduce a bandit environment to isolate and evaluate HT’s ability to infer optimal actions. This environment presents no temporal structure and requires the model to reason about uncertainty using only a static context of noisy action-reward pairs. The difficulty lies in the absence of exploration and in the challenge of identifying high-reward actions from sparse, biased samples. 

\subsubsection{Darkroom} The Darkroom dataset consists of 100,000 in-context datasets, each containing $H=100$ steps, generated using a uniformly random policy. These datasets are evenly distributed across 100 distinct goals. Query states are sampled uniformly from the state space, and optimal actions are determined by the following rule: the agent first moves vertically (up or down) until its $y-coordinate$ matches the goal's $y-coordinate$, then moves horizontally (left or right) until its $x-coordinate$ aligns with the goal. From this dataset, $80,000$ samples (corresponding to the first 80 goals) are used for training, while the remaining $20,000$ samples are reserved for validation.

\subsubsection{Darkroom Stochastic} This variant builds on the standard Darkroom environment, keeping the same $10 \times 10$ grid layout, goal distribution, and reward structure. The key difference lies in the transition dynamics: when the agent intends to move in one direction, there is a 20\%, 40\%, or 60\% probability (depending on the variant) that it instead moves left or right, simulating action misdirection. Not moving remains deterministic. This stochasticity introduces significant transition noise, requiring the model to learn robust strategies under uncertainty. As in the deterministic case, we generate 100,000 in-context datasets with $H = 100$ steps, collected using a uniformly random policy and split into 80,000 training and 20,000 validation examples. Optimal actions are computed based on the original goal-directed policy, ignoring stochasticity.

\subsubsection{Darkroom Hard} Similar to Darkroom, the Darkroom Hard dataset includes $100,000$ in-context datasets, each with $H=400$ steps, generated using a uniformly random policy. The key difference lies in the goal space: the dataset is evenly distributed across 1,600 distinct goals, corresponding to a $40\times40$ grid. Optimal actions are determined in the same way as in Darkroom. From the dataset, $80,000$ samples are used for training, associated with $1,280$ training goals, while $20,000$ samples are reserved for validation, associated with $320$ validation goals. Training and test goals are evenly distributed among the in-context datasets to ensure consistency, with no overlap between the goal sets.

\subsubsection{Miniworld} The Miniworld task uses image-based observations and involves only four distinct tasks, each corresponding to a specific colored box. Since no new tasks are introduced during testing, the number of in-context datasets required for pretraining is significantly smaller. Specifically, $40,000$ datasets are used, each consisting of $H=50$ steps. To reduce computational overhead, these datasets include only $(s,a,r)$ tuples. Query states consist of image observations and directional information, sampled uniformly across the state space. In this environment, the agent is placed at a random location, facing a random direction. Optimal actions are determined as follows: if the agent is not facing the correct box (within $\pm$15 degrees), it turns toward it; otherwise, it moves forward. Of the $40,000$ datasets, $32,000$ are used for training, while the remaining $8,000$ are reserved for validation.

\subsubsection{MuJoCo} 
\begin{figure}[t!]
  \centering

  \begin{subfigure}{0.48\linewidth}
    \centering
    \includegraphics[width=\linewidth]{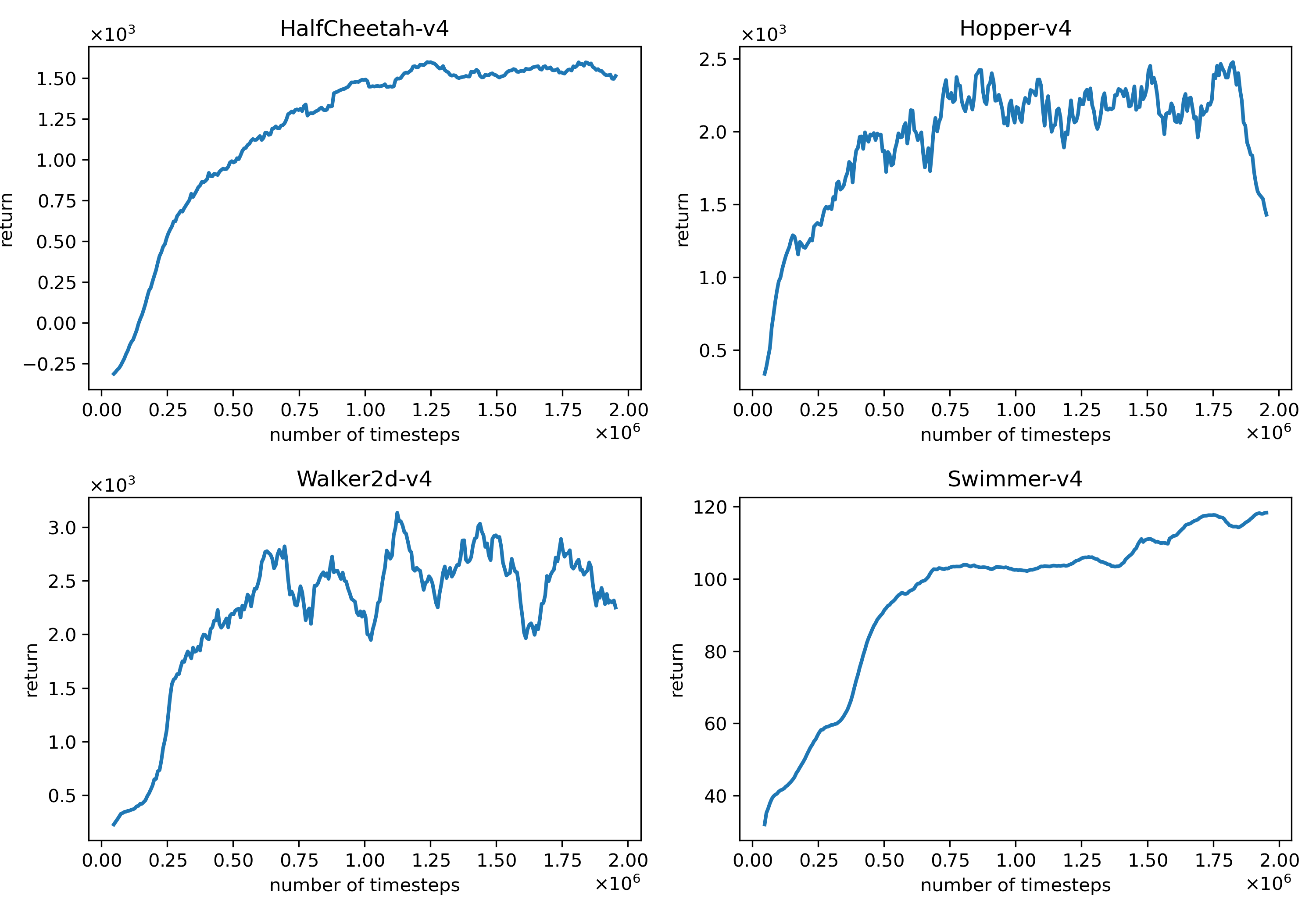}
    \caption{PPO}
    \label{fig:train_mujoco:ppo}
  \end{subfigure}\hfill
  \begin{subfigure}{0.48\linewidth}
    \centering
    \includegraphics[width=\linewidth]{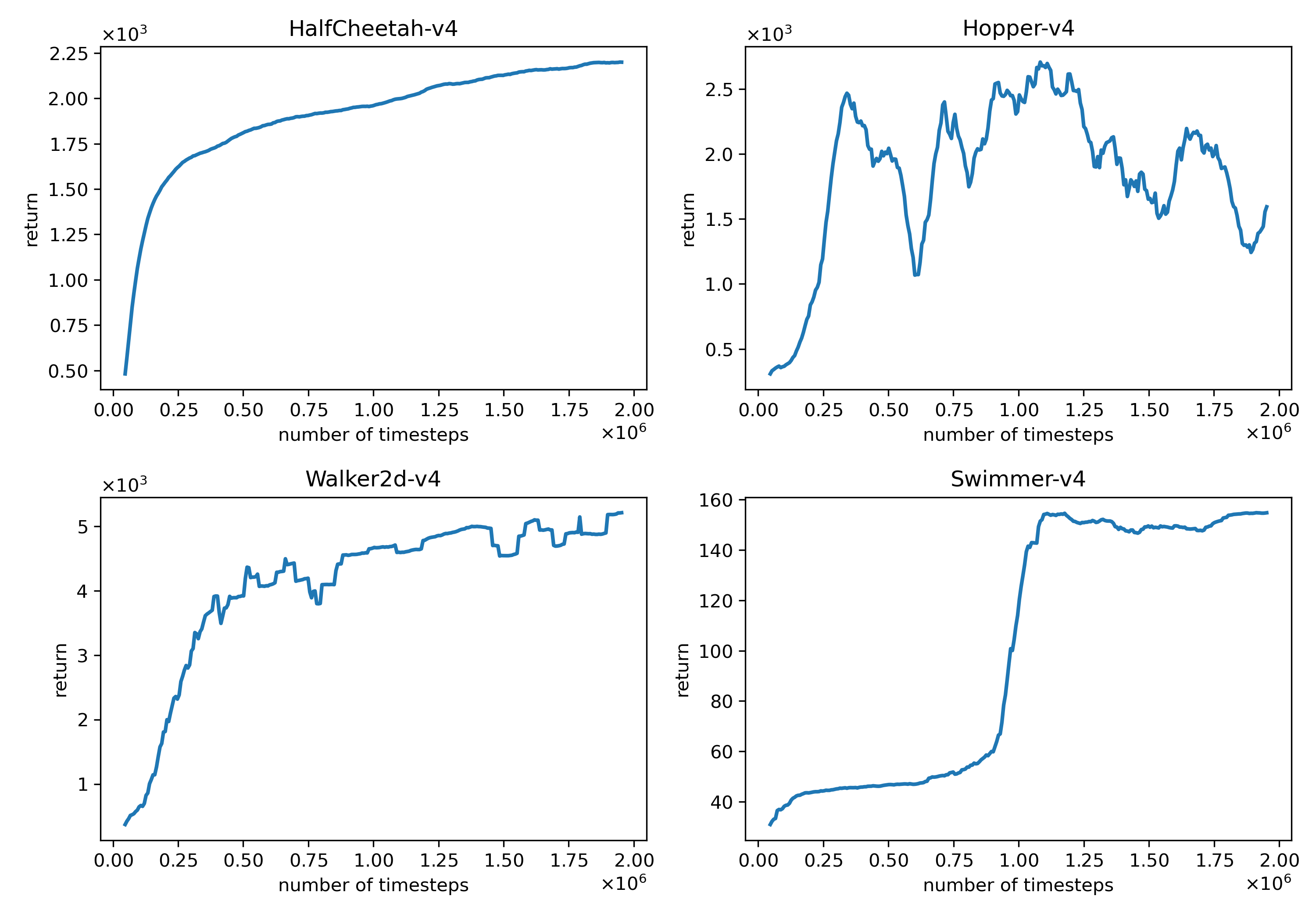}
    \caption{SAC}
    \label{fig:train_mujoco:sac}
  \end{subfigure}

  \caption{Training curves of PPO and SAC used to generate the training data in MuJoCo environments.}
  \label{fig:train_mujoco}
\end{figure}

For each MuJoCo environment—Hopper-v4, Walker2d-v4, HalfCheetah-v4, and Swimmer-v4—we generate 100,000 in-context datasets, each containing $H = 2000$ steps. These datasets are constructed from rollouts collected using expert policies trained with PPO, SAC, or a 50/50 mixture of both algorithms. Each dataset consists of $(s, a, r, s')$ transitions, with query states sampled uniformly from the respective environment’s state space. Optimal actions are defined as the action taken by the expert at that state. Table \ref{fig:train_mujoco} shows the training curves of PPO and SAC, which were finally used to generate the training datasets for the MuJoCo environments.

\subsubsection{Bandit} We generate 5-armed bandit tasks with rewards $R(\cdot | s, a) = \mathcal{N}(\mu_a, \sigma^2)$, where $\mu_a \sim \text{Unif}[0, 1]$ and $\sigma = 0.3$. Action frequencies in the pretraining datasets $D_{\text{pre}}$ are drawn by combining samples from a Dirichlet distribution with a point mass on a randomly chosen arm to promote diversity. The optimal policy for each task $\tau$ is computed as $\pi^*_\tau = \arg\max_a \mu_a$, and the model $M_\theta$ is trained to predict the optimal action from datasets of size up to $500$.


\subsection{SAC and PPO Training Configuration}

To generate the expert datasets for pretraining, we trained agents using the Proximal Policy Optimization (PPO) \citep{schulman2017proximal} and Soft Actor-Critic (SAC) \citep{haarnoja2018soft} algorithms on four standard MuJoCo environments: Hopper-v4, Walker2d-v4, HalfCheetah-v4, and Swimmer-v4. Training was conducted using Stable-Baselines3 in combination with Gymnasium for environment interfaces. For each environment and algorithm combination, agents were trained from scratch for 2 million environment steps using the default MlpPolicy architecture, which consists of fully connected layers. All training runs were seeded with a fixed random seed (42) to ensure reproducibility. Figure \ref{fig:train_mujoco} displays the training curves for each algorithm and environment used to generate the expert datasets for the MuJoCo domains. The models were trained with the default hyperparameters provided by Stable-Baselines3, which include an actor and critic learning rate of 3e-4, batch size 256 (SAC) or 64 (PPO), and discount factor $\gamma = 0.99$. SAC used automatic entropy tuning with a target entropy set to $-|\mathcal{A}|$, where $\mathcal{A}$ is the action space dimension.

\subsection{Experimental Setup for Multi-Armed Bandits}\label{app:bandit}

Additionally, we tested \acronym in the multi-armed bandit setting—a well-established and simplified instance of a MDP — where the state space is a singleton and the decision horizon is limited to a single time step ($H = 1$). We assess the performance of \acronym in the \textit{online} learning contexts.  The agent interacts with the environment to maximize cumulative reward, requiring a careful balance between exploration and exploitation to minimize regret.

\subsubsection{Baselines}

We compare \acronym against DPT and several widely studied bandit algorithms, each with a distinct approach to handling uncertainty:

\paragraph{Empirical Mean (Emp):}
This algorithm selects the action with the highest empirical mean reward and requires no hyperparameters. To avoid degenerate behavior, Emp is modified as follows: Each action is initially sampled once before the empirical strategy is applied. These adjustments improve its reliability.

\paragraph{Upper Confidence Bound (UCB):}
Based on Hoeffding’s inequality, UCB selects actions as
\[
\hat{a} \in \arg\max_{a \in \mathcal{A}} \hat{\mu}_a + \frac{1}{n_a}
\]
where $\hat{\mu}_a$ is the empirical mean reward and $n_a$ is the number of times action $a$ has been chosen. The confidence bonus constant was selected through coarse hyperparameter tuning.

\paragraph{Thompson Sampling (TS):}
Although the true reward means are sampled uniformly from $[0, 1]$, we employ a Gaussian model with prior mean $1/2$ and variance $1/12$ to approximate the true distribution. The noise model uses the correct variance. 

\paragraph{DPT:} The DPT model uses a transformer architecture with the following configuration: embedding size of 32, context length of 500, 4 hidden layers, and 2 attention heads per layer. Training uses the AdamW optimizer (learning rate $1 \times 10^{-4}$, weight decay $1 \times 10^{-4}$, batch size 64). For all experiments, the in-context dataset $D$ is randomly shuffled.

\subsubsection{Evaluation Metrics}

We assess performance with standard bandit evaluation metrics:

\begin{itemize}
   \item Cumulative regret, computed as $\sum_h (\mu_{a^*} - \mu_{a_h})$, where $a_k$ is the action chosen at time $h$.
\end{itemize}

\subsection{Pseudo-code: Training and Deployment}

\begin{algorithm}
\caption{\ourname (\acronym)}
\begin{algorithmic}[1]
\STATE \textbf{// Collecting pretraining dataset}
\STATE Initialize empty pretraining dataset $\mathcal{B}$
\FOR{$i \in [N]$}
    \STATE Sample task $\tau \sim \mathcal{T}_{\text{pre}}$, in-context dataset $\mathcal{D} \sim \mathcal{D}_{\text{pre}}(\cdot; \tau)$, query state $s_{\text{query}} \sim \mathcal{D}_{\text{query}}$
    \STATE Sample label $a^\star \sim \pi^\star_\tau(\cdot|s_{\text{query}})$ and add $(s_{\text{query}}, \mathcal{D}, a^\star)$ to $\mathcal{B}$
\ENDFOR
\STATE \textbf{(1) Train belief}
\STATE Initialize prior $p_\Phi$ and variational posterior $q_\phi$ with parameters $\Phi$ and $\phi$
\WHILE{not converged}
    \STATE Sample $(\mathcal{D})$ from $\mathcal{B}$ and predict $b_{j}(\cdot)=q_{\phi}(\cdot | D_{j})$ and $\hat{r}_{j}=p_{\Phi}(\cdot|m_j)$ for all $j \in [n]$
    \STATE Compute loss in \ref{eq:4} and update $\Phi$ and $\phi$
\ENDWHILE
\STATE \textbf{(2) Train policy model}
\STATE Initialize model $M_\theta$ with parameters $\theta$
\WHILE{not converged}
    \STATE Sample $(s_{\text{query}}, \mathcal{D}, a^\star)$ from $\mathcal{B}$
    \STATE Calculate the belief $b$ based on $\mathcal{D}$
    \STATE Predict $\hat{p}_j(\cdot) = M_\theta(\cdot| \mathcal{D}_j, b_{j}, s_{\text{query}})$ for all $j \in [n]$
    \STATE Compute loss in \ref{eq:5} with respect to $a^\star$ and update $\theta$
\ENDWHILE
\STATE \textbf{// Online test-time deployment}
\STATE Sample unknown task $\tau \sim \mathcal{T}_{\text{test}}$ and initialize empty $\mathcal{D} = \{\}$
\FOR{$\text{ep}$ in max eps}
    \STATE Deploy $M_\theta$ by sampling $a_h \sim M_\theta(\cdot|\mathcal{D}, b_h, s_h)$ at step $h$
    \STATE Add $(s_1, a_1, r_1, \ldots)$ to $\mathcal{D}$
\ENDFOR
\end{algorithmic}
\end{algorithm}

\subsection{Further Experimental \& Ablation Results}

We evaluate \acronym in the Bandit setting to test whether it can learn useful decision-making priors in the absence of temporal structure or long-term dependencies. In Bandit the agent is tasked with selecting one of five arms, each associated with an unknown reward distribution. The means of these distributions are randomly sampled from a uniform distribution, and the agent must infer the optimal arm based on a limited in-context dataset containing action-reward pairs. Bandit tasks offer minimal context—single-step episodes with independent rewards per arm — we set $H = 1$. This setting is deliberately challenging for a belief-based model like \acronym, as the short horizon limits the value of modeling uncertainty over reward distributions. Indeed, we do not expect HT to outperform specialized baselines here. Nevertheless, as shown in Figure~\ref{fig:ablation}(a), \acronym achieves comparable performance to classical exploration-based methods like UCB and Thompson Sampling (TS), while slightly outperforming DPT. Emp selects the arm with the highest empirical mean, and LCB favors underexplored arms with pessimistic estimates. In contrast, UCB and TS are designed for efficient exploration, which explains their strong performance. That \acronym matches these methods despite not being explicitly trained for exploration suggests it has learned a generalizable decision-making heuristic that remains effective even in simple, highly-optimized settings.


\subsubsection{Ablation Results}

\begin{figure}[h!]
  \centering

  \begin{subfigure}{0.3\linewidth}
    \centering
    \includegraphics[width=\linewidth]{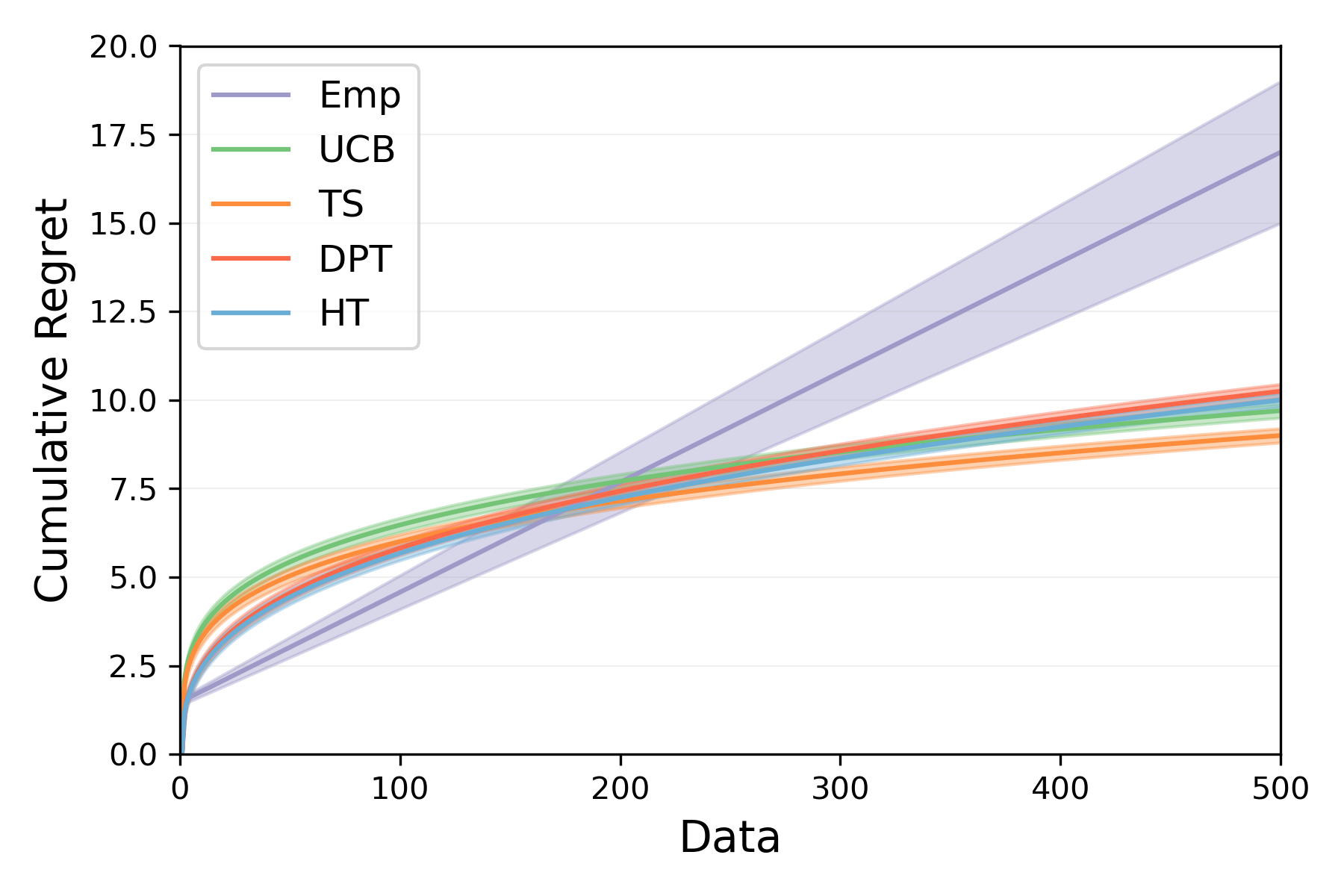}
    \caption{Bandit}
    \label{fig:ablation:bandit}
  \end{subfigure}\hfill
  \begin{subfigure}{0.3\linewidth}
    \centering
    \includegraphics[width=\linewidth]{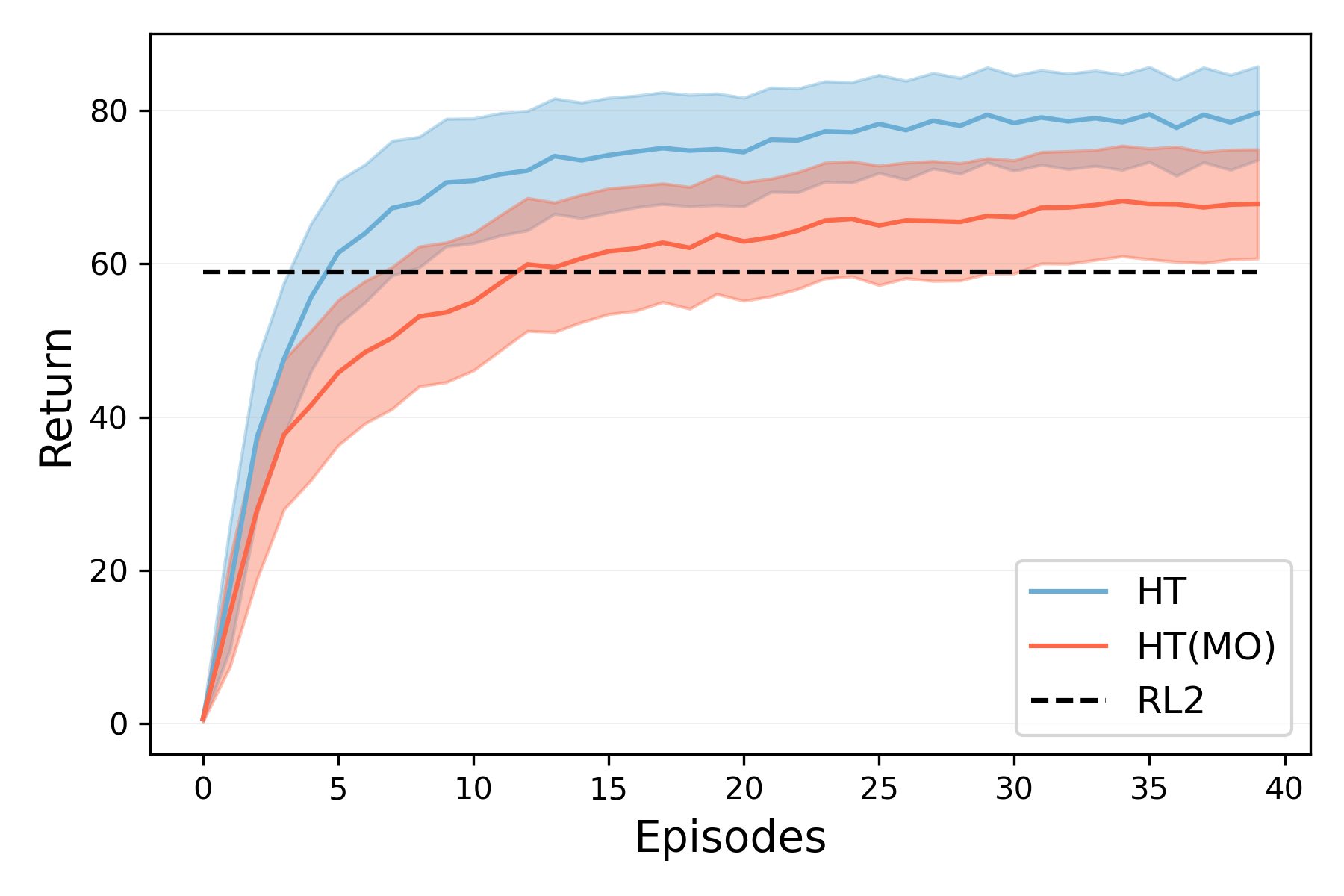}
    \caption{Darkroom}
    \label{fig:ablation:darkroom}
  \end{subfigure}\hfill
  \begin{subfigure}{0.3\linewidth}
    \centering
    \includegraphics[width=\linewidth]{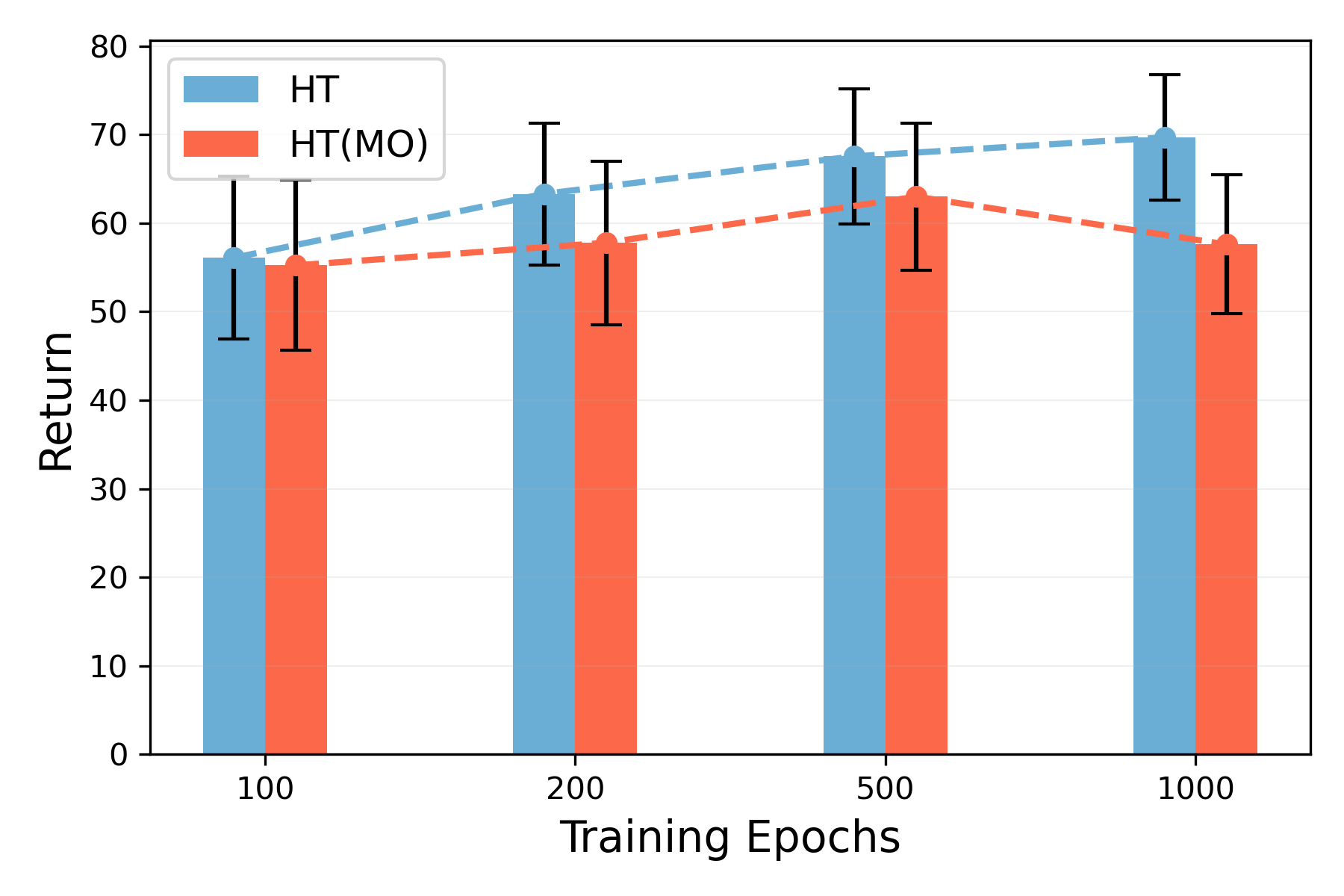}
    \caption{Darkroom (epochs)}
    \label{fig:ablation:darkroom-epoch}
  \end{subfigure}

  \caption{(a) Online cumulative regret in the Bandit environment. 
  (b) Online performance on test goals in Darkroom between HT and HT (MO). 
  (c) Online performance on test goals in Darkroom, after certain pre-training epochs between HT and HT (MO). 
  Reported are the mean return $\pm$ standard deviation on 20 trials across 10 seeds.}
  \label{fig:ablation}
\end{figure}

Incorporating a multi-objective loss function within the transformer framework is both theoretically sound and practically effective for modeling the belief distribution $b$. This approach leverages the transformer's ability to capture complex dependencies, enabling the seamless integration of Bayesian reasoning within the policy model $M_{\theta}$, thereby rendering the first pre-training phase redundant. By incorporating a multi-objective loss, the model can simultaneously maintain a belief distribution over rewards and refine its policy, effectively closing the loop between environment inference and optimal decision-making. This synergy enhances the model's capacity to generalize across tasks by capturing reward dynamics more effectively. As a result, this framework represents a principled and computationally efficient method for approximating what we term the Heuristic Bayesian Policy. In such a framework the loss function would be:

\begin{equation}\label{eq:ablation}
\begin{split}
    \mathcal{L}_{3}(\theta, \theta_{\text{action}},\theta_{\text{reward}}) = \\ \min_{\theta} \mathbb{E}_{P_{\text{pre}}} \sum_{j \in [n]} \Big[-\log\big( \lambda_{1} M_{\theta_{\text{action}}, \theta}(a^\ast | D_j, s_{\text{query}}) + \\ \lambda_{2} M_{\theta_{\text{reward}}, \theta}(r | D_j)\big) \Big],
\end{split}
\end{equation}

where $M_{\theta_{\text{action}}, \theta}$ and $M_{\theta_{\text{reward}}, \theta}$ are the action and reward prediction heads with their respective parameters and the shared parameters $\theta$, with $\lambda_{1}, \lambda_{2}$ being weighting parameters. We refer to this model as \ourname(MO). Figures \ref{fig:ablation}(b) and (c) show the mean episode return achieved by \ourname and \ourname(MO) on 20 trials across 10 seeds in Darkroom. While \ourname(MO) is theoretically plausible, empirical results do not show it performing on par with \ourname. Additional implementation details and hyperparameters are provided in the appendix.

\subsection{Hyperparameter Analysis}

In this section, we present \ourname's performance (mean ± standard deviation) across different configurations of the transformer's hyperparameters in Darkroom, specifically varying the number of attention heads and layers.

\begin{figure}[h]
  \centering

  \begin{subfigure}{0.45\linewidth}
    \centering
    \includegraphics[width=\linewidth]{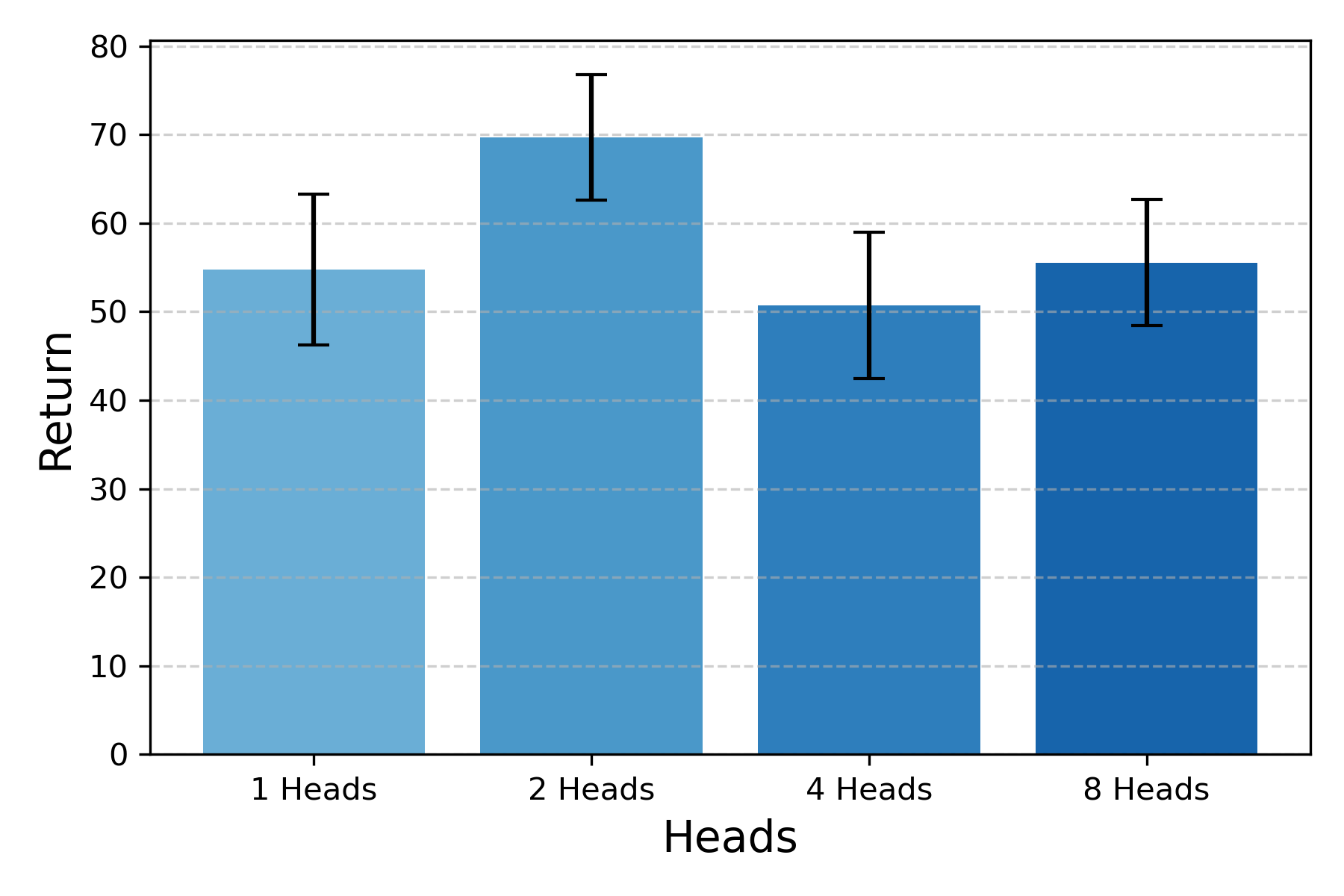}
    \caption{Attention Heads}
    \label{fig:heads}
  \end{subfigure}\hfill
  \begin{subfigure}{0.45\linewidth}
    \centering
    \includegraphics[width=\linewidth]{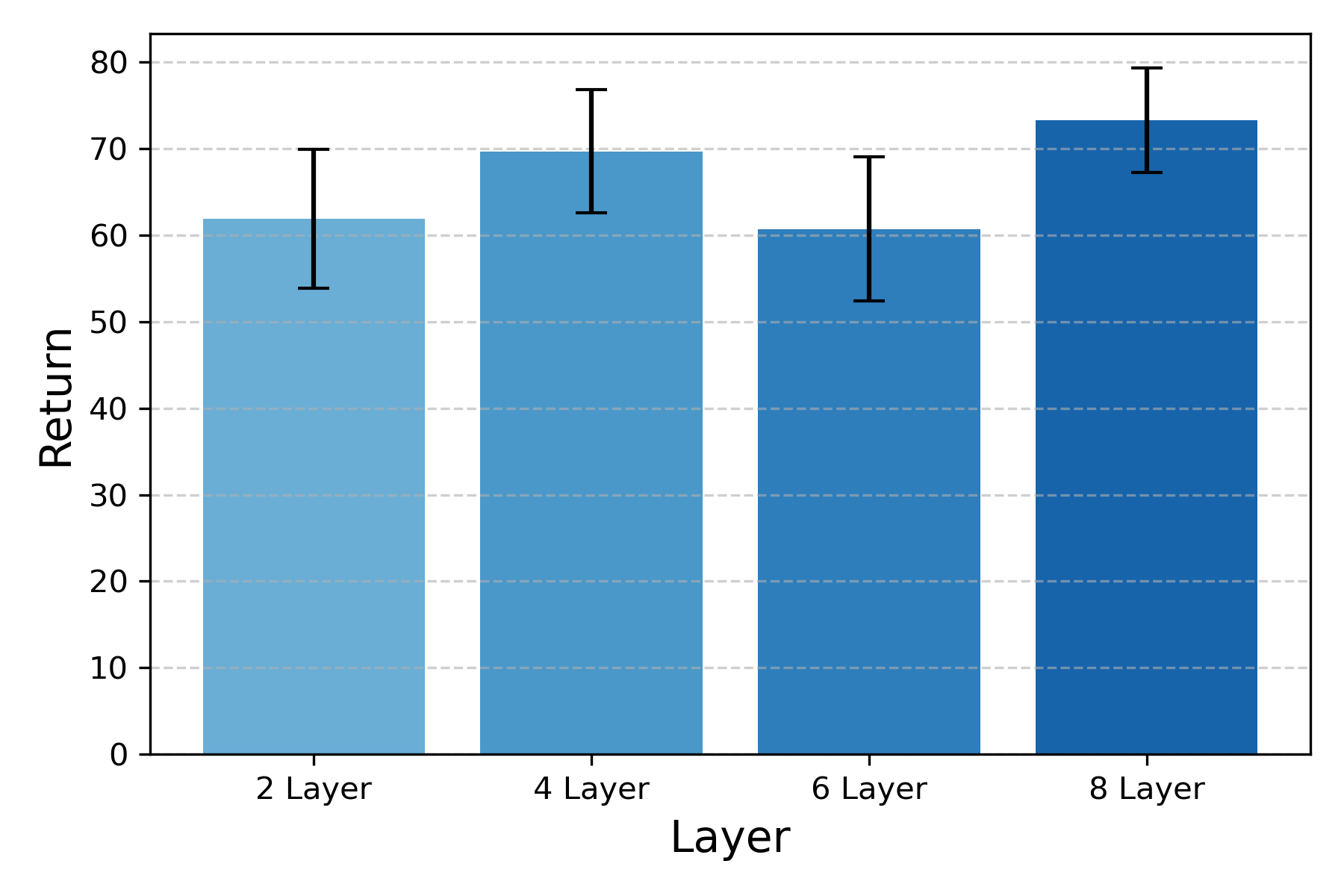}
    \caption{Layer}
    \label{fig:layers}
  \end{subfigure}

  \caption{Hyperparameter sensitivity for (a) the number of attention heads and (b) the number of layers.}
  \label{fig:hyperparams}
\end{figure}



\end{document}